\documentclass{article}
\usepackage{ijcai24}

\usepackage{times}
\usepackage{soul}
\usepackage{url}
\usepackage[hidelinks]{hyperref}
\usepackage[utf8]{inputenc}
\usepackage[small]{caption}
\usepackage{graphicx}
\usepackage{amsmath}
\usepackage{amsthm}
\usepackage{booktabs}
\usepackage{algorithm}
\usepackage{algorithmic}
\usepackage[switch]{lineno}
\usepackage{bm,amssymb,tabularx,tcolorbox,subcaption}

\title{Deep Reinforcement Learning with Hybrid Intrinsic Reward Model}

\author{
Mingqi Yuan$^1$
\and
Bo Li$^1$\and
Xin Jin$^{2}$\thanks{Corresponding author}\And
Wenjun Zeng$^2$\\
\affiliations
$^1$Department of Computing, The Hong Kong Polytechnic University, Hong Kong SAR, China\\
$^2$Ningbo Institute of Digital Twin, Eastern Institute of Technology, Ningbo, Zhejiang, China\\
\emails
mingqi.yuan@connect.polyu.hk,
comp-bo.li@polyu.edu.hk,
\{jinxin, wzeng\}@eitech.edu.cn
}

\begin{document}

\maketitle

\begin{abstract}
Intrinsic reward shaping has emerged as a prevalent approach to solving hard-exploration and sparse-rewards environments in reinforcement learning (RL). While single intrinsic rewards, such as curiosity-driven or novelty-based methods, have shown effectiveness, they often limit the diversity and efficiency of exploration. Moreover, the potential and principle of combining multiple intrinsic rewards remains insufficiently explored. To address this gap, we introduce \textbf{HIRE} (\textbf{H}ybrid \textbf{I}ntrinsic \textbf{RE}ward), a flexible and elegant framework for creating hybrid intrinsic rewards through deliberate fusion strategies. With HIRE, we conduct a systematic analysis of the application of hybrid intrinsic rewards in both general and unsupervised RL across multiple benchmarks. Extensive experiments demonstrate that HIRE can significantly enhance exploration efficiency and diversity, as well as skill acquisition in complex and dynamic settings. 
\end{abstract}

\section{Introduction}
Traditional reinforcement learning (RL) processes are fundamentally tied to extrinsic rewards, which are explicitly provided by the environment to incentivize specific goal-directed behaviors \cite{sutton2018reinforcement}. However, this approach often struggles in scenarios where extrinsic rewards are delayed, sparse, or entirely absent \cite{pathak2017curiosity}. Moreover, designing suitable extrinsic rewards for complex environments is consistently challenging, requiring substantial domain expertise. Poorly designed rewards can severely hinder the agents' learning efficiency and lead to suboptimal behavior. To overcome these limitations, intrinsic rewards have been introduced as auxiliary learning signals that motivate agents to engage in goal-independent behaviors, significantly enhancing their exploration and learning efficiency \cite{stadie2015incentivizing,bellemare2016unifying,pathak2017curiosity,ostrovski2017count,tang2017exploration,machado2020count,raileanu2020ride,yuan2022renyi}. For instance, \cite{burda2018exploration} proposed random network distillation (RND) that uses the prediction error against a fixed network as the intrinsic reward, encouraging the agent to visit those infrequently-seen states. \cite{seo2021state} suggested
maximizing the Shannon entropy of the state visitation distribution and proposed RE3, which utilizes a $k$-nearest neighbor estimator to make efficient entropy estimation and divides the sample mean into particle-based intrinsic rewards. RE3 can significantly promote the sample efficiency of model-based and model-free RL algorithms without any representation learning. However, these methods prevalently rely on single motivations, which limits their ability to address the diverse challenges present in complex and dynamic environments.


Natural agents (e.g., humans) often make decisions based on an interplay of biological, social, and cognitive motivations, as described by models of combined motivations like Maslow’s hierarchy of needs \cite{maslow1958dynamic} and existence-relatedness-growth (ERG) theory \cite{alderfer1972existence}. Inspired by this, hybrid intrinsic rewards have been proposed to provide agents with more comprehensive exploration incentives by combining multiple motivations. For example, NGU \cite{badia2020never} combines episodic and lifelong state novelty to generate intrinsic rewards. The episodic state novelty is evaluated using an episodic memory and pseudo-counts method, encouraging the agent to explore diverse states within each episode. Meanwhile, lifelong novelty is computed using RND, promoting exploration across episodes. NGU is the first algorithm to achieve non-zero rewards in the \textit{Pitfall!} game without using demonstrations or hand-crafted features. Similarly, RIDE \cite{raileanu2020ride} uses the difference between consecutive state embeddings as an intrinsic reward to encourage actions that cause significant state changes. To prevent agents from lingering between familiar states, RIDE discounts rewards based on episodic state visitation counts. Furthermore, \cite{henaff2023study} investigated the combination of global and episodic intrinsic rewards in contextual Markov decision processes (MDPs) and achieved a new state-of-the-art (SOTA) performance in the MiniHack benchmark. 

While the pioneering works mentioned above have achieved significant success, the full potential of combining multiple intrinsic motivations remains insufficiently explored. Current approaches typically rely on specific combinations of intrinsic rewards \cite{henaff2023study}, but they lack a systematic study and fail to provide generalizable principles for combining intrinsic rewards under different conditions. To address this gap, we introduce \textbf{HIRE}: \textbf{H}ybrid \textbf{I}ntrinsic \textbf{RE}ward framework that incorporates simple and efficient fusion strategies to blend diverse intrinsic rewards seamlessly. We summarize the contributions of this work as follows:
\begin{itemize}
    \item We developed a HIRE framework that includes four fusion strategies, two of which are newly proposed. HIRE is designed to support an arbitrary number of single intrinsic rewards and can be seamlessly integrated with a wide range of RL algorithms, providing a versatile tool for enhancing exploration in complex environments.
    
    \item We conducted an in-depth and systematic analysis of the application of hybrid intrinsic rewards in RL, focusing on the effects of various fusion strategies and the number of combined motivations. Specifically, we examined how different configurations (e.g., category and quantity) of multiple intrinsic rewards impact exploration diversity and efficiency. Extensive experiments were performed on recognized benchmarks, such as MiniGrid and Procgen, demonstrating the strengths and limitations of each configuration.

    \item We further examine hybrid intrinsic rewards on unsupervised RL tasks, encouraging agents to accumulate diverse experiences through a richer set of exploration incentives. Experimental results in the arcade learning environment (ALE) indicate that our approach significantly outperforms existing methods that rely on a single intrinsic reward, revealing the benefits of hybrid reward structures in unsupervised RL settings.
\end{itemize}

\begin{figure*}[t!]
    \centering
    \includegraphics[width=\linewidth]{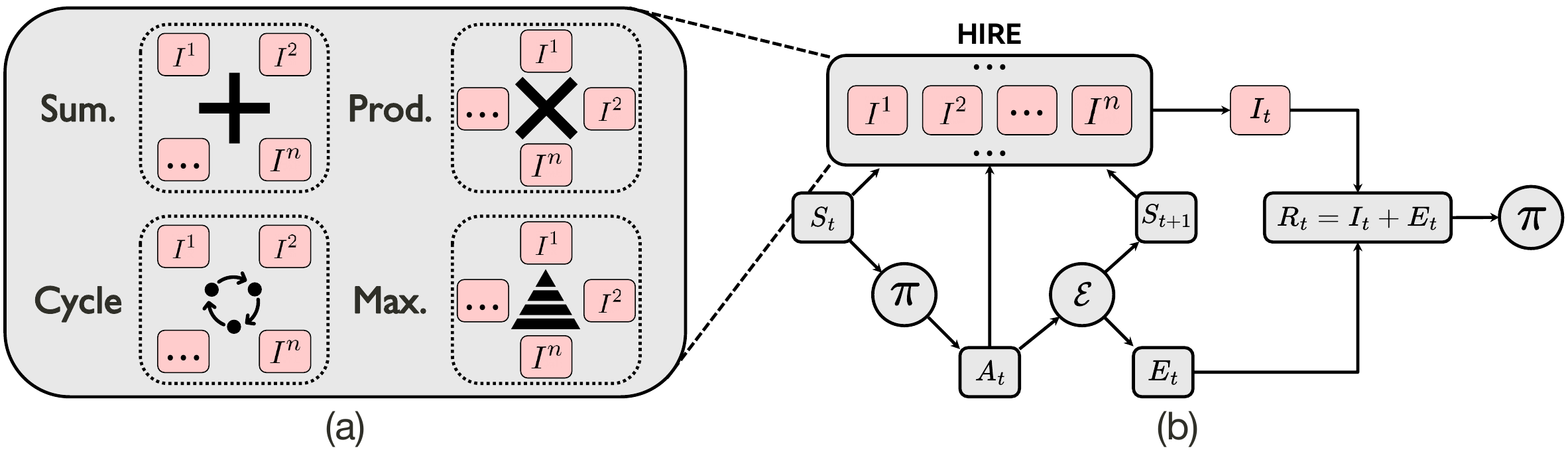}
    \caption{The overview of the HIRE framework. (a) Four reward fusion strategies implemented in HIRE. (b) HIRE is designed to be fully modular and decoupled from the RL training loop and can be integrated seamlessly with arbitrary RL algorithms.}
    \label{fig:workflow}
\end{figure*}

\section{Related Work}
\subsection{Intrinsic Reward Shaping}
Intrinsic reward shaping aims to encourage exploration by offering additional rewards to the RL agent based on its intrinsic learning motivation. These approaches can be broadly categorized into three main types: (i) count-based exploration \cite{bellemare2016unifying,burda2018exploration,hazan2019provably,seo2021state,yarats2021reinforcement,yuan2022rewarding,yuan2022renyi}, (ii) curiosity-driven exploration \cite{stadie2015incentivizing,pathak2017curiosity,pathak2019self,raileanu2020ride}, and (iii) skill discovery \cite{gregor2016variational,eysenbach2018diversity,liu2021aps,laskin2021cic,park2022lipschitz}. For example, \cite{pathak2017curiosity} designed the intrinsic curiosity module (ICM) to learn a joint embedding space with inverse and forward dynamics losses and was the first curiosity-based method successfully applied to deep RL settings. \cite{pathak2019self} further extended ICM by proposing Disagreement, which computes curiosity based on the variance among an ensemble of forward-dynamics models. Additionally, \cite{henaff2022exploration} introduced the E3B that generalizes count-based episodic bonuses to continuous state spaces. It encourages the exploration of diverse states within a learned embedding space for each episode. 

In this paper, we seek to establish a hybrid intrinsic reward framework that provides novel and efficient fusion strategies for combining diverse intrinsic rewards. We select ICM \cite{pathak2017curiosity}, NGU \cite{badia2020never}, RE3 \cite{seo2021state}, and E3B \cite{henaff2022exploration} as the candidates for our experiments, spanning the reward categories discussed above.

\subsection{Hybrid Intrinsic Reward}
As the RL community tackles increasingly complex problems, from singleton MDPs to contextual MDPs \cite{cobbe2020leveraging,samvelyan2021minihack}, hybrid intrinsic rewards have been introduced to provide more robust and comprehensive exploration incentives. A representative way is to combine global and episodic exploration bonuses \cite{badia2020never,raileanu2020ride,zhang2021noveld,mu2022improving}. For instance, \cite{flet-berliac2021adversarially} proposed AGAC that combines the Kullback–Leibler (KL) divergence between the behavior policy and adversary policy and episodic state visitation counts, which encourages the policy to adopt different behaviors as it tries to remain different from the adversary. \cite{zhang2021noveld} proposed NovelD that uses the difference between RND bonuses at two consecutive time steps, regulated by an episodic count-based bonus. \cite{mu2022improving} further explores the use of language as a general medium for highlighting relevant abstractions in an environment and extends NovelD using language abstractions. 

However, these methods often focus on limited types and quantities of intrinsic motivations, without exploring the impact of reward structure and failing to offer generalizable principles for their integration. In this paper, we further extend the boundary of hybrid intrinsic rewards by incorporating a broader array of distinct intrinsic rewards across both quantity and category levels. Our framework aims to enhance exploration robustness and enable RL agents to better adapt to complex and dynamic environments.

\subsection{Unsupervised RL}
Unsupervised reinforcement learning (URL) aims to pre-train agents without explicit supervision, enabling them to efficiently adapt to new tasks with minimal guidance \cite{laskin2020curl,campos2020explore,liu2021aps,yarats2021reinforcement}. Inspired by human learning, URL leverages intrinsic motivations to encourage exploration and skill acquisition in the absence of external rewards. The URL benchmark (URLB) \cite{laskin2021urlb} provides implementations of eight different URL algorithms and evaluates their performance using a modified version of the DeepMind Control Suite. However, these approaches only leverage single intrinsic motivations for pre-training.

In this paper, we make the pioneering attempt to apply hybrid intrinsic rewards in the context of URL. By introducing a richer, multi-motivational approach, our framework fosters diverse skill discovery and improves the effectiveness of pre-training.

\section{Background} \label{sec:background}
    We frame the RL problem considering a MDP \cite{bellman1957markovian,kaelbling1998planning} defined by a tuple $\mathcal{M}=(\mathcal{S},\mathcal{A},E,P,d_{0},\gamma)$, where $\mathcal{S}$ is the state space, $\mathcal{A}$ is the action space, and $E:\mathcal{S}\times\mathcal{A}\rightarrow\mathbb{R}$ is the extrinsic reward function, $P:\mathcal{S}\times\mathcal{A}\rightarrow\Delta(\mathcal{S})$ is the transition function that defines a probability distribution over $\mathcal{S}$, $d_{0}\in\Delta(\mathcal{S})$ is the distribution of the initial observation $\bm{s}_{0}$, and $\gamma\in[0, 1)$ is a discount factor. The goal of RL is to learn a policy $\pi_{\bm\theta}(\bm{a}|\bm{s})$ to maximize the expected discounted return:
	\begin{equation}
		J_{\pi}(\bm{\theta})=\mathbb{E}_{\pi}\left[\sum_{t=0}^{\infty}\gamma^{t}E_{t}\right].
	\end{equation}

 Furthermore, letting $\mathcal{I}=\{I^{i}\}_{i=1}^{n}$ denote a set of single intrinsic reward functions, where $I^{i}:\mathcal{S}\times\mathcal{A}\rightarrow\mathbb{R}$ represents a specific intrinsic motivation signal. To unify these signals, we introduce a hybrid reward model $f:\mathbb{R}^{n}\rightarrow\mathbb{R}$, which combines multiple intrinsic rewards. The resulting augmented optimization objective becomes
	\begin{equation}\label{eq:additive}
		J_{\pi}(\bm{\theta})=\mathbb{E}_{\pi}\left[\sum_{t=0}^{\infty}\gamma^{t}\bigg(E_{t}+\beta_{t}\cdot f(\mathcal{I})\bigg)\right],
	\end{equation}
 where $\beta_{t}=\beta_{0}(1-\kappa)^{t}$ controls the degree of exploration, and $\kappa$ is a decay rate.

\section{Hybrid Intrinsic Reward Framework}
\subsection{Architecture}
In this section, we propose HIRE, a flexible framework that offers four efficient fusion strategies for constructing hybrid intrinsic rewards in RL, namely \textbf{summation}, \textbf{product}, \textbf{cycle}, and \textbf{maximum}, respectively. The formulation of each strategy is described in Table~\ref{tb:fusion methods}. As shown in Figure~\ref{fig:workflow}, HIRE is designed to be fully modular and decoupled from the RL training loop, allowing it to integrate seamlessly with any RL algorithm. Moreover, HIRE supports the combination of any number and type of single intrinsic reward. To isolate the effects of intrinsic rewards, we adopt a simple additive model where intrinsic and extrinsic rewards are combined linearly, as defined in Eq.~(\ref{eq:additive}). This approach ensures that the influence of intrinsic rewards on exploration can be effectively evaluated without interference from complex reward structures.

\begin{table}[t!]
\centering
\begin{tabular}{ll}
\toprule[1.0pt]
\textbf{Strategy} & \textbf{Formulation}   \\ \midrule[1.0pt]
Summation (S)     & $I_{t} = \sum_{i=1}^{n} w^{i}_{t}\cdot I^{i}_{t}$         \\ \midrule
Product (P)       & $I_{t} = \prod_{i=1}^{n} I^{i}_{t}$                \\ \midrule
Cycle (C)         & $I_{t} = I^{i}_{t}, i=(t\mod n)$        \\ \midrule
Maximum (M)       & $I_{t} = \max \{I_{t}^{i}\}_{i=1}^{n}$  \\ \bottomrule[1.0pt]
\end{tabular}
\caption{Formulations of the four implemented fusion strategies.}
\label{tb:fusion methods}
\end{table}

\subsection{Fusion Strategy Analysis}
We analyze the potential advantages and limitations associated with each strategy as follows.

\textbf{Summation (S)}. The summation strategy combines intrinsic rewards linearly, with each reward $I^{i}$ weighted by a coefficient $w^{i}$. It is straightforward to implement and flexible, enabling the agent to utilize multiple intrinsic
motivations simultaneously for broader exploration. \textit{However, its effectiveness hinges on carefully balanced weights, as improper tuning can lead to skewed exploration and conflicting signals, which may reduce exploration efficiency.}

\textbf{Product (P)}. The product strategy incorporates intrinsic rewards with a multiplicative approach, adopted by multiple methods \cite{badia2020never,raileanu2020ride,henaff2022exploration,zhang2021noveld}, such as NGU \cite{badia2020never}, which utilizes a product of lifelong and episodic state novelty. It forces the agent to satisfy multiple motivations simultaneously and leads to well-rounded exploration. \textit{However, it is highly sensitive to low reward values, as any near-zero signal can collapse the overall product, making it less stable in environments with fluctuating rewards.}

Based on the summation and product strategies, we further propose two new fusion strategies: \textbf{Cycle} and \textbf{Maximum}. 

\textbf{Cycle (C)}. The cycle strategy combines the extrinsic reward with one intrinsic reward at a time, cycling through them across time steps. By iteratively focusing on different motivations, it ensures all intrinsic rewards are utilized and reduces the reliance on any single reward type. This robustness can enhance the agent’s ability to adapt to changing environments and challenges, as it fosters a broader understanding of the task dynamics. \textit{This dynamic approach also allows the agent to avoid the pitfalls of reward imbalance and conflicting signals, offering a more stable and adaptive exploration process.}


\textbf{Maximum (M)}. The maximum strategy selects the highest intrinsic reward at each time step, emphasizing the most significant motivation at any moment. It mimics human learning, where individuals often prefer tasks or topics that provide the most immediate satisfaction or engagement. \textit{By prioritizing the most salient reward, this strategy ensures efficient exploration and rapid adaptation to novel environments, while minimizing the risk of being misled by less relevant signals.}


The cycle and maximum strategies can be viewed as special cases of the summation method, where only one non-zero weight exists at a time. Equipped with these four strategies, HIRE provides an elegant framework for creating hybrid intrinsic rewards tailored to various exploration needs. Finally, to simplify the notation, the generated hybrid intrinsic rewards are denoted by \textbf{HIRE-\{Type\}\{$n$\}}. For example, \textbf{HIRE-S2} represents the summation of two intrinsic rewards, and \textbf{HIRE-P3} represents the product of three intrinsic rewards.

\section{Experiments}
In this section, we design the experiments to achieve the two main objectives: (i) evaluate the performance of the HIRE framework on challenging tasks, and (ii) conduct a systematic analysis of the application of hybrid intrinsic rewards. 

\subsection{Experimental Settings}

\begin{figure}[t!]
    \centering
    \includegraphics[width=\linewidth]{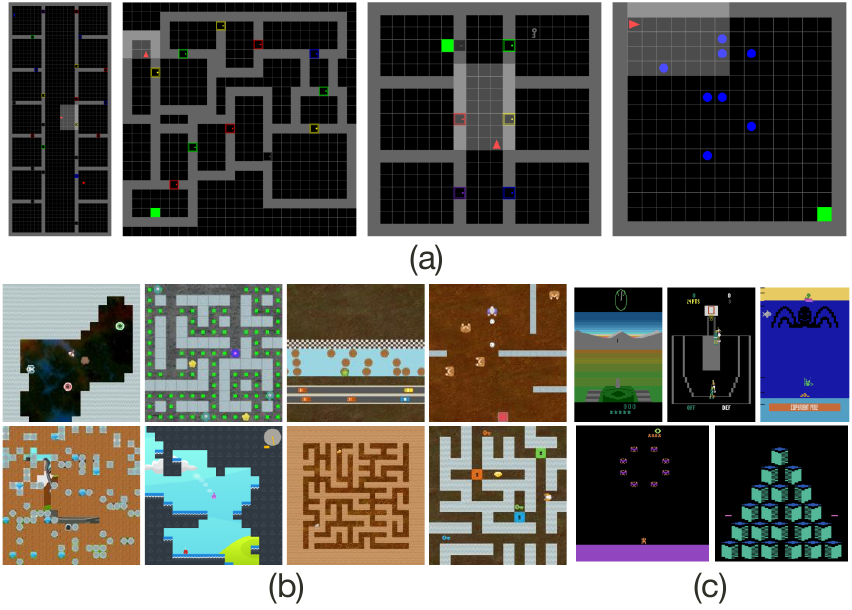}
    \caption{Screenshots of the experiment environments. (a) From left to right: \textit{KeyCorridorS10R7}, \textit{MultiRoom-N12-S10}, \textit{LockedRoom}, and \textit{Dynamic-Obstacles-16×16}. (b) Eight navigation and exploration environments from the \textit{Procgen} benchmark. (c) \textit{ALE-5}.}
    \label{fig:screenshots}
\end{figure}

We first conduct a series of experiments on the MiniGrid \cite{MinigridMiniworld23} and Procgen \cite{cobbe2020leveraging} benchmarks. MiniGrid is a collection of 2D grid-world environments with goal-oriented tasks, which can effectively examine agents' exploration capabilities by presenting challenging exploration and sparse-rewards scenarios. Previous studies have also highlighted the effectiveness of intrinsic rewards in MiniGrid environments \cite{raileanu2020ride,henaff2022exploration,henaff2023study}. In contrast, Procgen presents a more diverse set of challenges with visually rich and dynamically changing environments that require robust exploration and adaptive behaviors. For each benchmark, we select eight hard-exploration and navigation tasks. Specifically, we have \textit{KeyCorridorS8R5}, \textit{KeyCorridorS9R6}, \textit{KeyCorridorS10R7}, \textit{MultiRoom-N7-S8}, \textit{MultiRoom-N10-S10}, \textit{MultiRoom-N12-S10}, \textit{LockedRoom}, and \textit{Dynamic-Obstacles-16×16} from MiniGrid, and \textit{CaveFlyer}, \textit{Chaser}, \textit{Dodgeball}, \textit{Heist}, \textit{Jumper}, \textit{Maze}, \textit{Miner}, and \textit{Plunder} from Procgen. The screenshots of these selected environments are shown in Figure~\ref{fig:screenshots}.

For the intrinsic reward set, we select ICM \cite{pathak2017curiosity}, NGU \cite{badia2020never}, RE3 \cite{seo2021state}, and E3B \cite{henaff2022exploration}. This set is designed to span a wide spectrum of intrinsic reward designs, such as curiosity-driven, count-based, and memory-based exploration. The formulation and implementation details of these selected intrinsic rewards can be found in Appendix~\ref{appendix:baselines} and Appendix~\ref{appendix:exp setting}. Equipped with the reward set, we design hybrid intrinsic rewards by traversing the combinations of these single intrinsic rewards and applying the four fusion strategies. For example, Table~\ref{tb:sum candidates} illustrates all the candidates from \textbf{HIRE-S0} to \textbf{HIRE-S4}. Similarly, we have the same combinations for all the other three fusion strategies.

\begin{table}[h!]
\centering
\small
\begin{tabular}{ll}
\toprule[1.0pt]
\textbf{Type} & \textbf{Candidates}                                                                        \\ \midrule[1.0pt]
HIRE-S0       & Extrinsic \\ \midrule
HIRE-S1       & ICM, NGU, RE3, E3B                                                                            \\ \midrule
HIRE-S2       & \begin{tabular}[c]{@{}l@{}}S(NGU, ICM), S(NGU, RE3), S(NGU, E3B)\\ S(E3B, RE3), S(E3B, ICM), S(RE3, ICM)\end{tabular} \\ \midrule
HIRE-S3       & \begin{tabular}[c]{@{}l@{}}S(NGU, E3B, RE3), S(NGU, RE3, ICM)\\ S(NGU, E3B, ICM), S(E3B, RE3, ICM)\end{tabular}   \\ \midrule
HIRE-S4       & S(NGU, E3B, RE3, ICM)                                                                               \\ \bottomrule[1.0pt]
\end{tabular}
\caption{All the reward candidates of the summation fusion strategy. These combinations also apply to the other three fusion strategies.}
\label{tb:sum candidates}
\end{table}

For the backbone RL algorithm, we select proximal policy optimization (PPO) \cite{schulman2017proximal} as the baseline. Importantly, as shown in Figure \ref{fig:workflow}, we keep the PPO hyperparameters fixed and the overall RL training loop unmodified throughout all the experiments to isolate the effect of intrinsic rewards. The fixed PPO hyperparameters are shown in Table~\ref{tb:ppo_params}.

\subsection{Results Analysis}
To demonstrate the results analysis more explicitly, we formulate a series of research questions and answer them in sequence.
\begin{figure*}[t!]
\centering
\begin{subfigure}[b]{\linewidth}
    \centering
    \includegraphics[width=\linewidth]{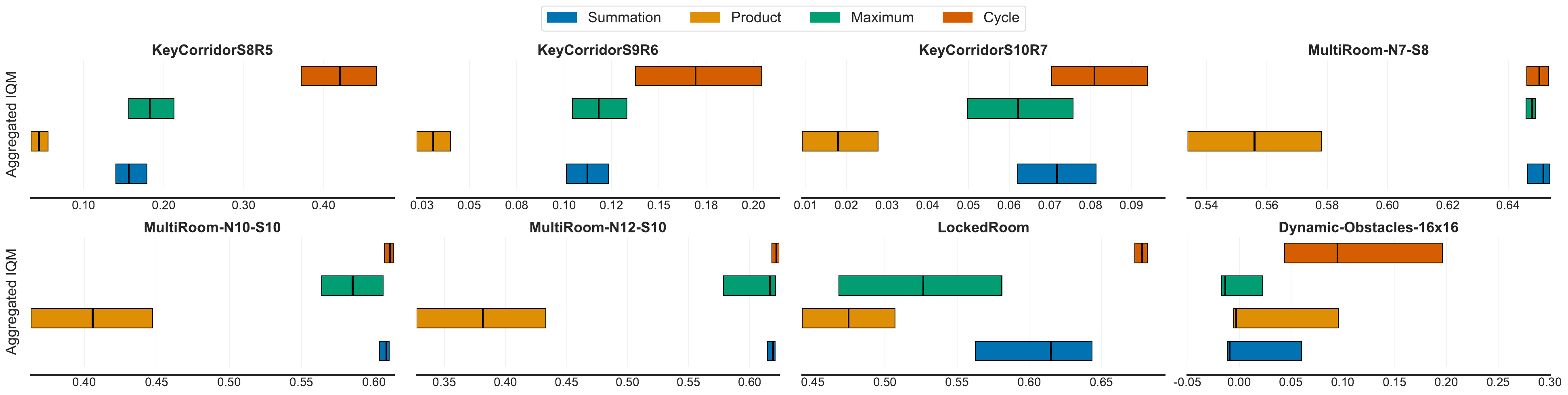}
    \caption{\textit{MiniGrid}}
\end{subfigure}
\begin{subfigure}[b]{\linewidth}
    \centering
    \includegraphics[width=\linewidth]{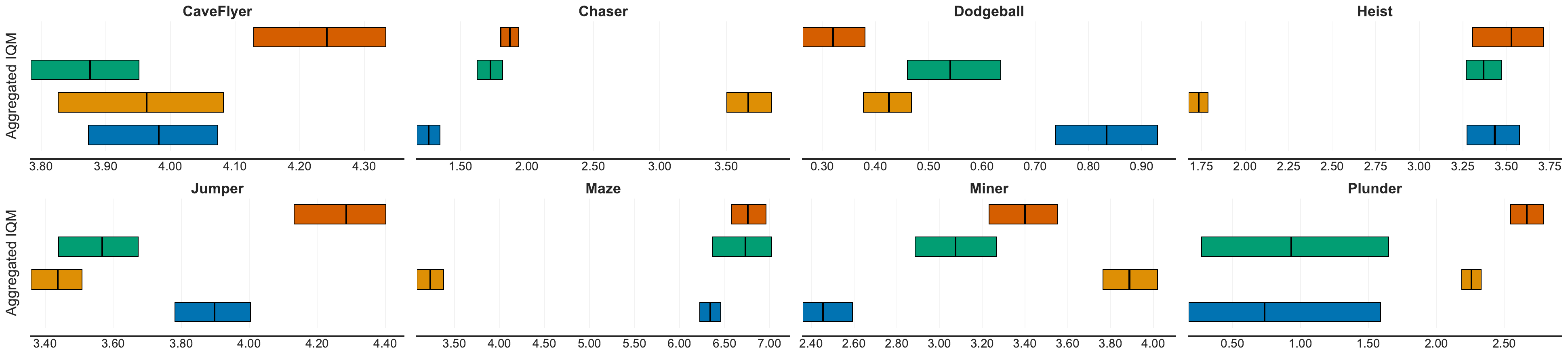}
    \caption{\textit{Procgen}}
\end{subfigure}
\caption{Strategy-level performance comparison on the MiniGrid and Procgen benchmarks. Here, each strategy corresponds to eleven reward candidates listed in Table~\ref{tb:sum candidates}. Bars indicate $95\%$ confidence intervals computed using stratified bootstrapping over five random seeds.}
\label{fig:minigrid_procgen_strategy_level}
\end{figure*}

\begin{center}
    \begin{tcolorbox}[colback=gray!10,
        colframe=black,
        width=\linewidth,
        arc=1mm, auto outer arc,
        boxrule=0.5pt,
        ]
        \textbf{Q1: Which fusion strategy is the most robust for hybrid intrinsic rewards?}
    \end{tcolorbox}
\end{center}

We begin with the analysis of the performance of each fusion strategy. Figure~\ref{fig:minigrid_procgen_strategy_level} illustrates the strategy-level performance comparison on the sixteen environments from MiniGrid and Procgen, in which the aggregated interquartile mean (IQM) is utilized as the key performance indicator (KPI) \cite{agarwal2021deep}. Overall, the cycle strategy demonstrates superior
robustness and achieves the best performance on most tasks. By periodically prioritizing different motivations, the cycle strategy enables the agent to adapt dynamically and balance exploration effectively. In contrast, the maximum and summation strategies achieve moderate and task-dependent performance in the two benchmarks. While the summation strategy provides relatively stable exploration, it lacks the adaptability required for dynamic environments where conflict signals may arise as the environment changes. Similarly, the maximum strategy that prioritizes the dominant intrinsic reward struggles to generalize across tasks due to its limited exploration diversity. Its greedy nature may be misled by inappropriate motivations and over-explore certain areas. These limitations were particularly evident in environments like \textit{Dynamic-Obstacles-16×16} and \textit{Plunder}, where broader and more adaptive exploration is required. The product strategy performs relatively poorly on the MiniGrid benchmark, especially for the \textit{KeyCorridor} and \textit{MultiRoom} environments where sequential tasks need to be addressed. However, it outperforms the summation and maximum strategies in the \textit{Dynamic-Obstacles-16×16} and excels in \textit{Chaser} and \textit{Miner}. This may be caused by its ability to amplify the synergy between multiple intrinsic motivations, enabling the agent to navigate the dynamic environment more effectively by prioritizing states that satisfy multiple exploration incentives. 

\begin{center}
    \begin{tcolorbox}[colback=gray!10,
        colframe=black,
        width=\linewidth,
        arc=1mm, auto outer arc,
        boxrule=0.5pt,
        ]
        \textbf{Q2: Which hybrid intrinsic reward is the best for each environment?}
    \end{tcolorbox}
\end{center}

\begin{figure*}[t!]
\centering
\begin{subfigure}[b]{\linewidth}
    \centering
    \includegraphics[width=\linewidth]{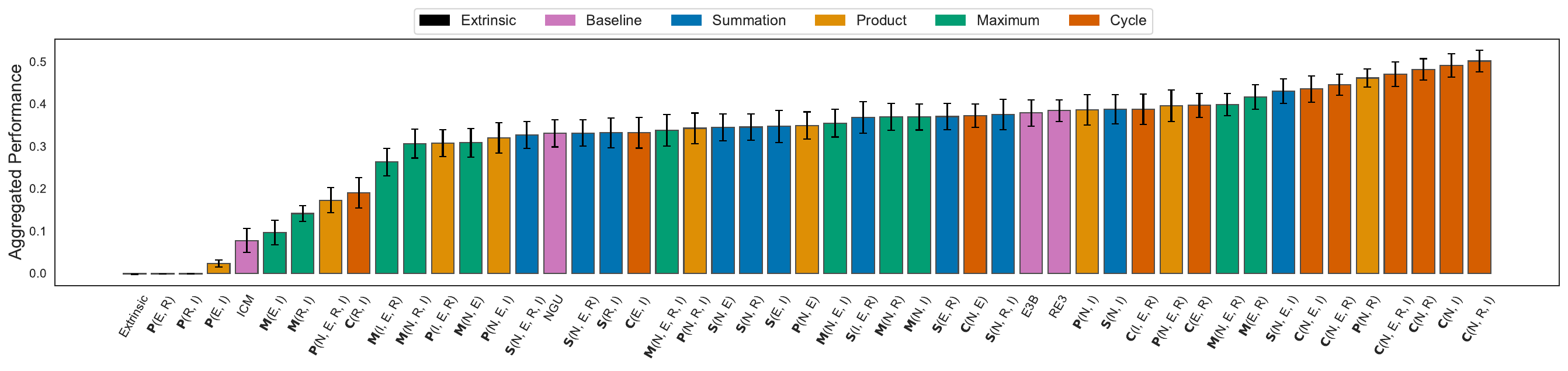}
\end{subfigure}
\begin{subfigure}[b]{\linewidth}
    \centering
    \includegraphics[width=\linewidth]{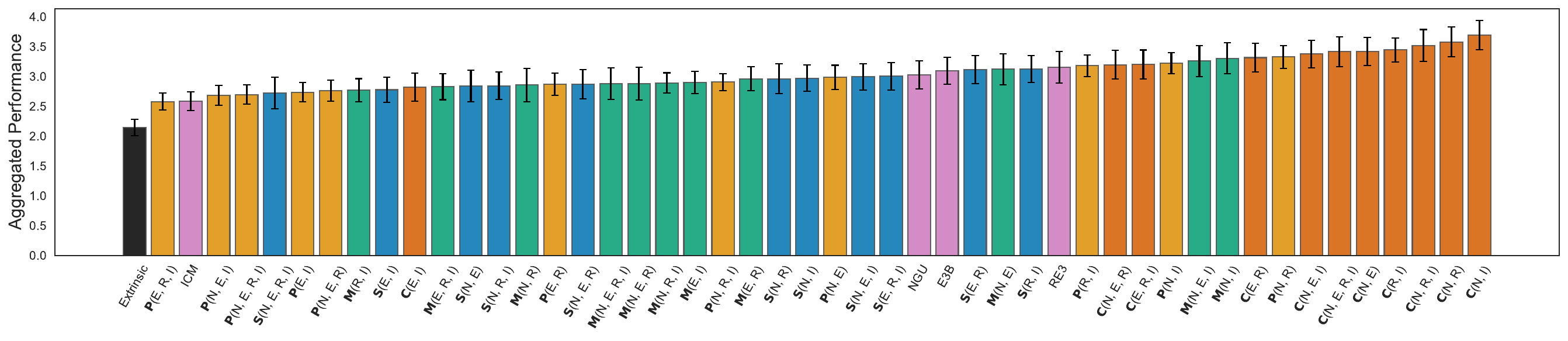}
\end{subfigure}
\caption{Aggregated performance ranking of all the reward candidates on the MiniGrid (top) and Procgen (bottom) benchmarks. For simplicity, we abbreviate \textbf{ICM}, \textbf{NGU}, \textbf{RE3}, and \textbf{E3B} as \textbf{I}, \textbf{N}, \textbf{R}, and \textbf{E}. The mean and standard error are computed across all the environments.}
\label{fig:minigrid_procgen_rank}
\end{figure*}

\begin{figure*}[h!]
\centering
\begin{subfigure}[b]{0.49\textwidth}
    \centering
    \includegraphics[width=\linewidth]{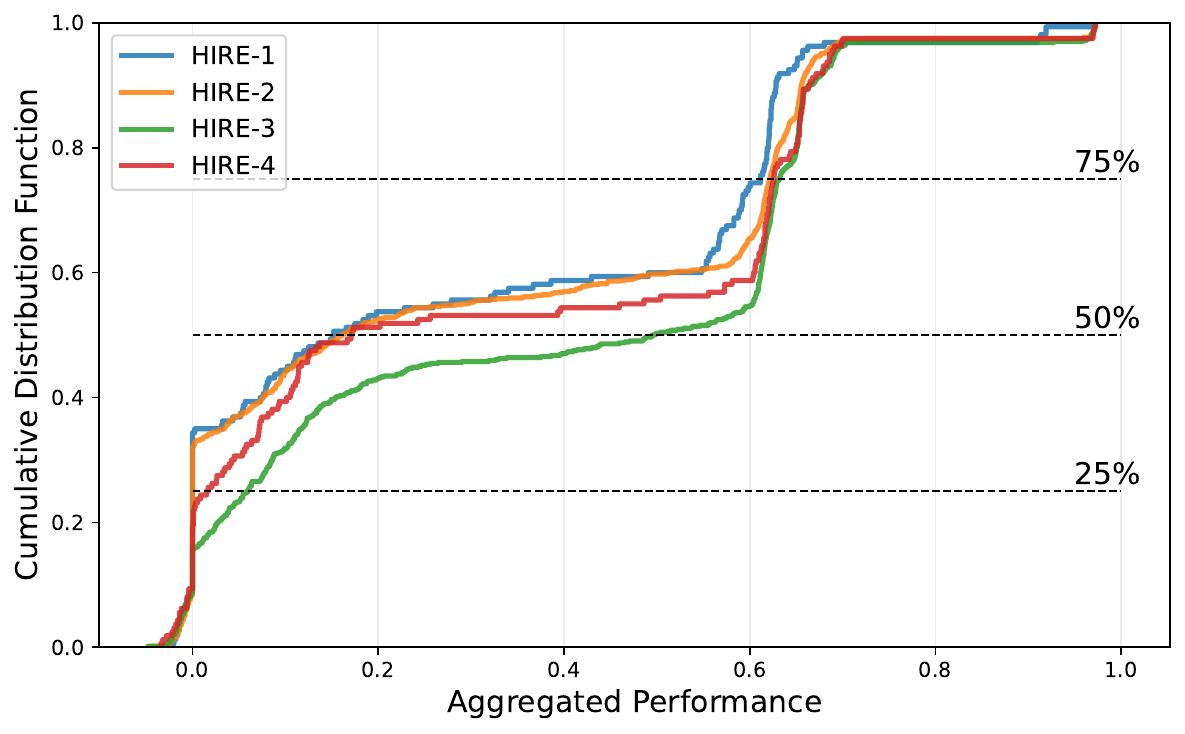}
\end{subfigure}
\hfill
\begin{subfigure}[b]{0.49\textwidth}
    \centering
    \includegraphics[width=\linewidth]{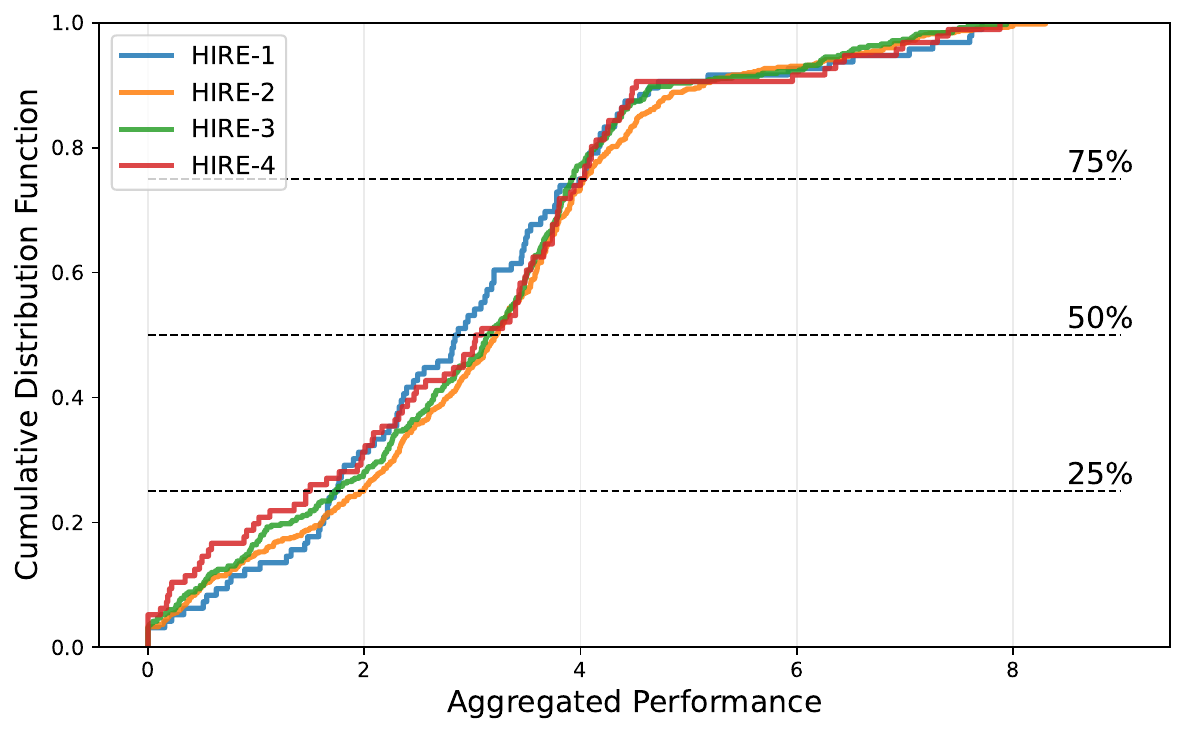}
\end{subfigure}
\caption{Cumulative distribution function of the performance from HIRE-1 to HIRE-4 on the MiniGrid (left) and Procgen (right) benchmarks.}
\label{fig:minigrid_number_level_cdf}
\end{figure*}

Next, we analyze the performance of each hybrid intrinsic reward candidate. We provide detailed performance rankings of all the candidates across all the experiment environments in Appendix~\ref{appendix:ranking}, and Table~\ref{tb:best_mg_procgen} lists the best reward candidate for each environment. Furthermore, Figure~\ref{fig:minigrid_procgen_rank} presents an aggregated performance ranking of all reward candidates, which suggests that \textbf{C}(NGU, RE3, ICM) and \textbf{C}(NGU, ICM) are the generally best reward candidates for MiniGrid and Procgen. Specifically, for MiniGrid, the candidates that utilize the cycle strategy achieved the highest performance in six environments, and the maximum and product strategies excel in one environment each. For Procgen, the cycle strategy ranks first in four environments, the product strategy wins two environments, and the maximum and summation strategies excel in one environment each.

\begin{center}
    \begin{tcolorbox}[colback=gray!10,
        colframe=black,
        width=\linewidth,
        arc=1mm, auto outer arc,
        boxrule=0.5pt,
        ]
        \textbf{Q3: Which single intrinsic rewards and combinations contribute the most?}
    \end{tcolorbox}
\end{center}

As shown in Figure~\ref{fig:minigrid_procgen_rank} and Table~\ref{tb:best_mg_procgen}, NGU contributes to twelve out of the sixteen best reward candidates, and RE3, E3B, and ICM contribute to ten, six, and nine candidates, respectively. NGU includes both global and episodic exploration bonuses, which offer comprehensive incentives for exploration, making it adaptable to a wide range of tasks. On the other hand, RE3 effectively promotes exploration without using auxiliary representation learning, allowing it to function well alongside other integrated intrinsic rewards. In particular, the (NGU, RE3) combination achieves the best performance in four environments, while the (NGU, RE3, ICM) combination demonstrates significant scores regarding both individual and overall performance. Based on the analysis above, we recommend the (NGU, RE3) as the best combination, which combines comprehensiveness of exploration and computational efficiency.

\begin{center}
    \begin{tcolorbox}[colback=gray!10,
        colframe=black,
        width=\linewidth,
        arc=1mm, auto outer arc,
        boxrule=0.5pt,
        ]
        \textbf{Q4: Does the performance of hybrid intrinsic rewards scale with the number of integrated single intrinsic rewards?}
    \end{tcolorbox}
\end{center}

\begin{figure*}[t!]
    \centering
    \includegraphics[width=\linewidth]{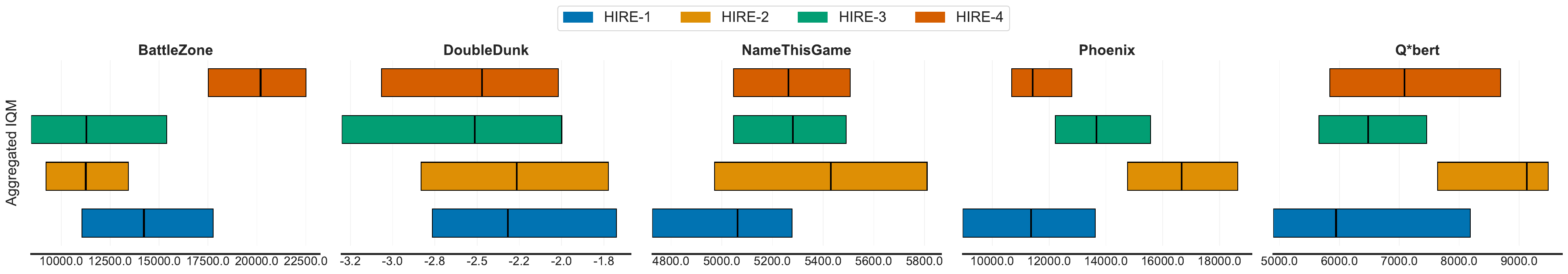}
    \caption{Quantity-level performance comparison on the ALE-5 benchmark. Here, each strategy corresponds to four reward candidates. The training is divided into the pre-training phase (intrinsic rewards only) and the fine-tuning phase (extrinsic rewards only), and each phase has five million environment steps. Bars indicate 95\% confidence intervals computed using stratified bootstrapping over five random seeds.}
    \label{fig:ale_all_algos}
\end{figure*}

Next, we conduct the performance comparison among the combinations of different numbers of single intrinsic rewards to investigate the quantity effect. Figure~\ref{fig:minigrid_number_level_single} and Figure~\ref{fig:procgen_number_level_single} illustrate the quantity-level performance comparison of each strategy in each environment. For MiniGrid, it is natural to find that the cycle and maximum strategies produce significant performance gains across environments as the number of rewards increases. The summation and product strategies do not benefit from the quantity effect explicitly, especially in the task with dynamic layouts. In contrast, for Procgen environments, the quantity effect is limited and degenerates the performance in environments like \textit{Dodgeball} and \textit{Chaser}. This indicates that balancing multiple exploration motivations is challenging in procedurally-generated environments. Figure~\ref{fig:minigrid_number_level_cdf} computes the cumulative distribution function (CDF) of the aggregated performance from HIRE-1 to HIRE-4, which indicates the three-reward combinations tend to perform better in MiniGrid environments, whereas two-reward combinations are generally more effective in Procgen environments. This analysis demonstrates that the quantity effect of hybrid intrinsic rewards is finite, especially in environments with high dynamics where too many rewards can lead to confusion in exploration priorities and suboptimal behavior.

\begin{figure}[h!]
    \centering
    \includegraphics[width=\linewidth]{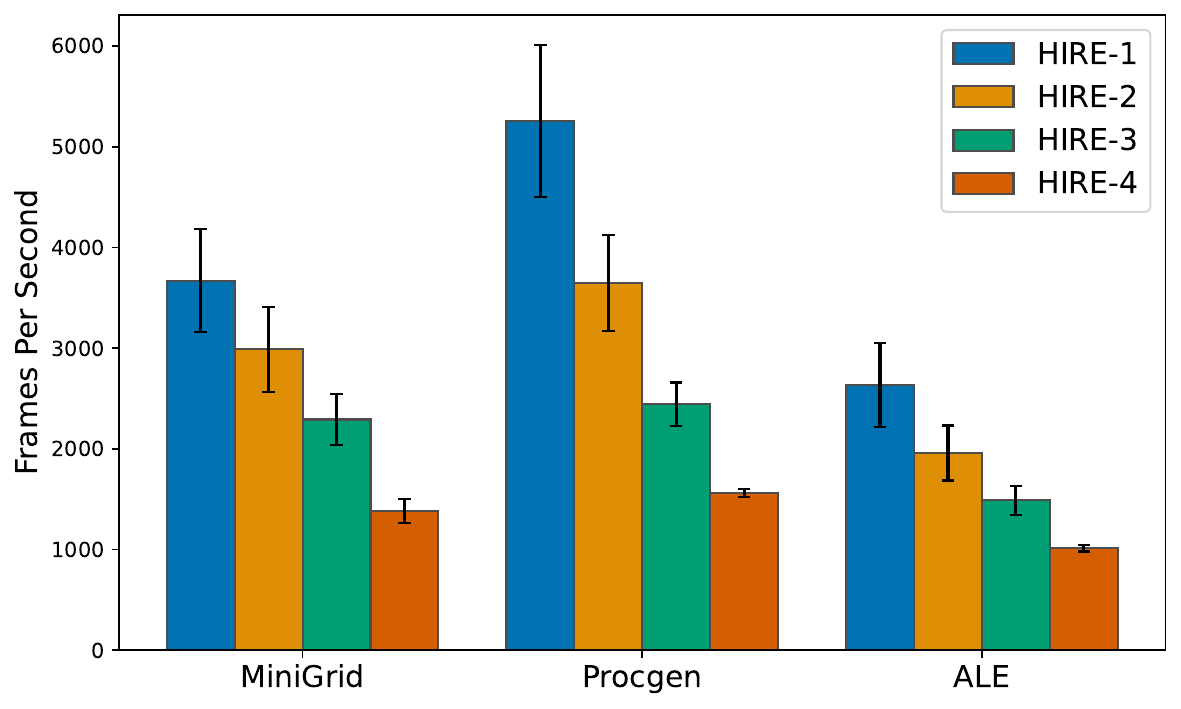}
    \caption{Computational efficiency from HIRE-1 to HIRE-4 on the three experiment benchmarks. All the test is performed using an AMD 7950X CPU and an NVIDIA RTX4090 GPU.}
    \label{fig:fps}
    \vskip -0.1in
\end{figure}

\begin{center}
    \begin{tcolorbox}[colback=gray!10,
        colframe=black,
        width=\linewidth,
        arc=1mm, auto outer arc,
        boxrule=0.5pt,
        ]
        \textbf{Q5: Can hybrid intrinsic rewards improve unsupervised RL performance compared to single intrinsic rewards?}
    \end{tcolorbox}
\end{center}

Furthermore, we evaluate the effectiveness of hybrid intrinsic rewards on unsupervised RL tasks using the arcade learning environment (ALE) benchmark \cite{bellemare2013arcade}. Specifically, we focus on a subset of ALE known as ALE-5, which includes the games \textit{BattleZone}, \textit{DoubleDunk}, \textit{NameThisGame}, \textit{Phoenix}, and \textit{Q*bert}. Research has shown that ALE-5 typically produces median score estimates for these 57 games that are within 10\% of their true values \cite{aitchison2023atari}. For reward candidates, we select the best-performing combinations based on the MiniGrid and Procgen experiments. Specifically, (NGU, RE3) and (NGU, RE3, ICM) are selected for HIRE-2 and HIRE-3, and they are tested with all four fusion strategies.

Figure~\ref{fig:ale_all_algos} illustrates the quantity-level performance comparison of selected reward candidates, and Table~\ref{tb:best_ale} lists the best candidate for each environment. The hybrid intrinsic rewards produce significant performance gains as compared to the single intrinsic reward approaches. Notably, both the cycle and maximum strategies excel in two environments. These results highlight the ability of hybrid rewards to encourage diverse skill discovery during the pre-training phase, leading to improved adaptation in downstream tasks.

\begin{center}
    \begin{tcolorbox}[colback=gray!10,
        colframe=black,
        width=\linewidth,
        arc=1mm, auto outer arc,
        boxrule=0.5pt,
        ]
        \textbf{Q6: How compute-efficient are the hybrid intrinsic rewards?}
    \end{tcolorbox}
\end{center}

Finally, we report the computation efficiency of different levels of hybrid intrinsic rewards. To make a fair comparison, we utilize the training frames per second (FPS) as the KPI. Figure~\ref{fig:fps} indicates that the training FPS decreases significantly as more rewards are integrated. These results suggest that HIRE configurations with up to three rewards strike a balance between exploration performance and computational cost.



\section{Discussion}
In this paper, we introduced the HIRE framework that incorporates four efficient fusion strategies for creating hybrid intrinsic rewards in an elegant manner. HIRE is highly modular and supports any type and number of single intrinsic rewards, which can be combined with arbitrary RL algorithms. We evaluate HIRE on multiple benchmarks (e.g., MiniGrid and Procgen) and conduct an in-depth and systematic study of the application of hybrid intrinsic rewards. Over 4000 experiments demonstrate that HIRE can significantly promote the RL agent's learning capabilities while revealing the strategy-level and quantity-level properties of the hybrid intrinsic rewards. Our findings aim to provide clear guidance for future research in intrinsically motivated RL.

Still, there are currently remaining limitations to this work. In our experiments, we selected four representative single intrinsic rewards to serve as the baseline. However, this reward set cannot encompass all the existing exploration algorithms, e.g., the skill-based algorithms like VISR \cite{Hansen2020Fast} and APS \cite{liu2021aps}. On the other hand, restricted by computational resources, it is difficult to investigate larger reward candidates like HIRE-5 or HIRE-6 further. Finally, we aim to evaluate HIRE in more real-world scenarios (e.g., robotics) to increase its applicability. These limitations will be addressed in future work.

\clearpage\newpage
\bibliographystyle{named}
\bibliography{ijcai24}

\begin{thebibliography}{}

\bibitem[\protect\citeauthoryear{Agarwal \bgroup \em et al.\egroup }{2021}]{agarwal2021deep}
Rishabh Agarwal, Max Schwarzer, Pablo~Samuel Castro, Aaron~C Courville, and Marc Bellemare.
\newblock Deep reinforcement learning at the edge of the statistical precipice.
\newblock {\em Advances in neural information processing systems}, 34:29304--29320, 2021.

\bibitem[\protect\citeauthoryear{Aitchison \bgroup \em et al.\egroup }{2023}]{aitchison2023atari}
Matthew Aitchison, Penny Sweetser, and Marcus Hutter.
\newblock Atari-5: Distilling the arcade learning environment down to five games.
\newblock In {\em International Conference on Machine Learning}, pages 421--438. PMLR, 2023.

\bibitem[\protect\citeauthoryear{Alderfer}{1972}]{alderfer1972existence}
Clayton~P Alderfer.
\newblock Existence, relatedness, and growth: Human needs in organizational settings.
\newblock {\em The Free Press google schola}, 2:1--39, 1972.

\bibitem[\protect\citeauthoryear{Auer}{2002}]{auer2002using}
Peter Auer.
\newblock Using confidence bounds for exploitation-exploration trade-offs.
\newblock {\em Journal of Machine Learning Research}, 3(Nov):397--422, 2002.

\bibitem[\protect\citeauthoryear{Badia \bgroup \em et al.\egroup }{2020}]{badia2020never}
Adri{\`a}~Puigdom{\`e}nech Badia, Pablo Sprechmann, Alex Vitvitskyi, Daniel Guo, Bilal Piot, Steven Kapturowski, Olivier Tieleman, Martin Arjovsky, Alexander Pritzel, Andrew Bolt, and Charles Blundell.
\newblock Never give up: Learning directed exploration strategies.
\newblock In {\em International Conference on Learning Representations}, 2020.

\bibitem[\protect\citeauthoryear{Bellemare \bgroup \em et al.\egroup }{2013}]{bellemare2013arcade}
Marc~G Bellemare, Yavar Naddaf, Joel Veness, and Michael Bowling.
\newblock The arcade learning environment: An evaluation platform for general agents.
\newblock {\em Journal of Artificial Intelligence Research}, 47:253--279, 2013.

\bibitem[\protect\citeauthoryear{Bellemare \bgroup \em et al.\egroup }{2016}]{bellemare2016unifying}
Marc Bellemare, Sriram Srinivasan, Georg Ostrovski, Tom Schaul, David Saxton, and Remi Munos.
\newblock Unifying count-based exploration and intrinsic motivation.
\newblock {\em Proceedings of Advances in Neural Information Processing Systems}, 29:1471--1479, 2016.

\bibitem[\protect\citeauthoryear{Bellman}{1957}]{bellman1957markovian}
Richard Bellman.
\newblock A markovian decision process.
\newblock {\em Journal of mathematics and mechanics}, pages 679--684, 1957.

\bibitem[\protect\citeauthoryear{Burda \bgroup \em et al.\egroup }{2019}]{burda2018exploration}
Yuri Burda, Harrison Edwards, Amos Storkey, and Oleg Klimov.
\newblock Exploration by random network distillation.
\newblock {\em Proceedings of the 7th International Conference on Learning Representations}, pages 1--17, 2019.

\bibitem[\protect\citeauthoryear{Campos \bgroup \em et al.\egroup }{2020}]{campos2020explore}
V{\'\i}ctor Campos, Alexander Trott, Caiming Xiong, Richard Socher, Xavier Gir{\'o}-i Nieto, and Jordi Torres.
\newblock Explore, discover and learn: Unsupervised discovery of state-covering skills.
\newblock In {\em International Conference on Machine Learning}, pages 1317--1327. PMLR, 2020.

\bibitem[\protect\citeauthoryear{Chevalier{-}Boisvert \bgroup \em et al.\egroup }{2023}]{MinigridMiniworld23}
Maxime Chevalier{-}Boisvert, Bolun Dai, Mark Towers, Rodrigo Perez{-}Vicente, Lucas Willems, Salem Lahlou, Suman Pal, Pablo~Samuel Castro, and Jordan Terry.
\newblock Minigrid {\&} miniworld: Modular {\&} customizable reinforcement learning environments for goal-oriented tasks.
\newblock In {\em Advances in Neural Information Processing Systems 36, New Orleans, LA, USA}, December 2023.

\bibitem[\protect\citeauthoryear{Cobbe \bgroup \em et al.\egroup }{2020}]{cobbe2020leveraging}
Karl Cobbe, Chris Hesse, Jacob Hilton, and John Schulman.
\newblock Leveraging procedural generation to benchmark reinforcement learning.
\newblock In {\em International conference on machine learning}, pages 2048--2056. PMLR, 2020.

\bibitem[\protect\citeauthoryear{Dani \bgroup \em et al.\egroup }{2008}]{dani2008stochastic}
Varsha Dani, Thomas~P Hayes, and Sham~M Kakade.
\newblock Stochastic linear optimization under bandit feedback.
\newblock In {\em COLT}, volume~2, page~3, 2008.

\bibitem[\protect\citeauthoryear{Eysenbach \bgroup \em et al.\egroup }{2018}]{eysenbach2018diversity}
Benjamin Eysenbach, Abhishek Gupta, Julian Ibarz, and Sergey Levine.
\newblock Diversity is all you need: Learning skills without a reward function.
\newblock In {\em International Conference on Learning Representations}, 2018.

\bibitem[\protect\citeauthoryear{Flet-Berliac \bgroup \em et al.\egroup }{2021}]{flet-berliac2021adversarially}
Yannis Flet-Berliac, Johan Ferret, Olivier Pietquin, Philippe Preux, and Matthieu Geist.
\newblock Adversarially guided actor-critic.
\newblock In {\em International Conference on Learning Representations}, 2021.

\bibitem[\protect\citeauthoryear{Gregor \bgroup \em et al.\egroup }{2016}]{gregor2016variational}
Karol Gregor, Danilo~Jimenez Rezende, and Daan Wierstra.
\newblock Variational intrinsic control.
\newblock {\em arXiv preprint arXiv:1611.07507}, 2016.

\bibitem[\protect\citeauthoryear{Hansen \bgroup \em et al.\egroup }{2020}]{Hansen2020Fast}
Steven Hansen, Will Dabney, Andre Barreto, David Warde-Farley, Tom~Van de~Wiele, and Volodymyr Mnih.
\newblock Fast task inference with variational intrinsic successor features.
\newblock In {\em International Conference on Learning Representations}, 2020.

\bibitem[\protect\citeauthoryear{Hazan \bgroup \em et al.\egroup }{2019}]{hazan2019provably}
Elad Hazan, Sham Kakade, Karan Singh, and Abby Van~Soest.
\newblock Provably efficient maximum entropy exploration.
\newblock In {\em Proceedings of the International Conference on Machine Learning}, pages 2681--2691, 2019.

\bibitem[\protect\citeauthoryear{Henaff \bgroup \em et al.\egroup }{2022}]{henaff2022exploration}
Mikael Henaff, Roberta Raileanu, Minqi Jiang, and Tim Rockt{\"a}schel.
\newblock Exploration via elliptical episodic bonuses.
\newblock {\em Advances in Neural Information Processing Systems}, 35:37631--37646, 2022.

\bibitem[\protect\citeauthoryear{Henaff \bgroup \em et al.\egroup }{2023}]{henaff2023study}
Mikael Henaff, Minqi Jiang, and Roberta Raileanu.
\newblock A study of global and episodic bonuses for exploration in contextual mdps.
\newblock {\em arXiv preprint arXiv:2306.03236}, 2023.

\bibitem[\protect\citeauthoryear{Kaelbling \bgroup \em et al.\egroup }{1998}]{kaelbling1998planning}
Leslie~Pack Kaelbling, Michael~L Littman, and Anthony~R Cassandra.
\newblock Planning and acting in partially observable stochastic domains.
\newblock {\em Artificial intelligence}, 101(1-2):99--134, 1998.

\bibitem[\protect\citeauthoryear{Laskin \bgroup \em et al.\egroup }{2020}]{laskin2020curl}
Michael Laskin, Aravind Srinivas, and Pieter Abbeel.
\newblock Curl: Contrastive unsupervised representations for reinforcement learning.
\newblock In {\em International conference on machine learning}, pages 5639--5650. PMLR, 2020.

\bibitem[\protect\citeauthoryear{Laskin \bgroup \em et al.\egroup }{2021a}]{laskin2021cic}
Michael Laskin, Hao Liu, Xue~Bin Peng, Denis Yarats, Aravind Rajeswaran, and Pieter Abbeel.
\newblock Cic: Contrastive intrinsic control for unsupervised skill discovery.
\newblock In {\em Deep RL Workshop NeurIPS 2021}, 2021.

\bibitem[\protect\citeauthoryear{Laskin \bgroup \em et al.\egroup }{2021b}]{laskin2021urlb}
Misha Laskin, Denis Yarats, Hao Liu, Kimin Lee, Albert Zhan, Kevin Lu, Catherine Cang, Lerrel Pinto, and Pieter Abbeel.
\newblock Urlb: Unsupervised reinforcement learning benchmark.
\newblock In J.~Vanschoren and S.~Yeung, editors, {\em Proceedings of the Neural Information Processing Systems Track on Datasets and Benchmarks}, volume~1, 2021.

\bibitem[\protect\citeauthoryear{Li \bgroup \em et al.\egroup }{2010}]{li2010contextual}
Lihong Li, Wei Chu, John Langford, and Robert~E Schapire.
\newblock A contextual-bandit approach to personalized news article recommendation.
\newblock In {\em Proceedings of the 19th international conference on World wide web}, pages 661--670, 2010.

\bibitem[\protect\citeauthoryear{Liu and Abbeel}{2021}]{liu2021aps}
Hao Liu and Pieter Abbeel.
\newblock Aps: Active pretraining with successor features.
\newblock In {\em International Conference on Machine Learning}, pages 6736--6747. PMLR, 2021.

\bibitem[\protect\citeauthoryear{Machado \bgroup \em et al.\egroup }{2020}]{machado2020count}
Marlos~C Machado, Marc~G Bellemare, and Michael Bowling.
\newblock Count-based exploration with the successor representation.
\newblock In {\em Proceedings of the AAAI Conference on Artificial Intelligence}, volume~34, pages 5125--5133, 2020.

\bibitem[\protect\citeauthoryear{Maslow}{1958}]{maslow1958dynamic}
Abraham~H Maslow.
\newblock A dynamic theory of human motivation.
\newblock 1958.

\bibitem[\protect\citeauthoryear{Mu \bgroup \em et al.\egroup }{2022}]{mu2022improving}
Jesse Mu, Victor Zhong, Roberta Raileanu, Minqi Jiang, Noah Goodman, Tim Rockt{\"a}schel, and Edward Grefenstette.
\newblock Improving intrinsic exploration with language abstractions.
\newblock {\em Advances in Neural Information Processing Systems}, 35:33947--33960, 2022.

\bibitem[\protect\citeauthoryear{Ostrovski \bgroup \em et al.\egroup }{2017}]{ostrovski2017count}
Georg Ostrovski, Marc~G Bellemare, A{\"a}ron Oord, and R{\'e}mi Munos.
\newblock Count-based exploration with neural density models.
\newblock In {\em Proceedings of the International Conference on Machine Learning}, pages 2721--2730, 2017.

\bibitem[\protect\citeauthoryear{Park \bgroup \em et al.\egroup }{2022}]{park2022lipschitz}
Seohong Park, Jongwook Choi, Jaekyeom Kim, Honglak Lee, and Gunhee Kim.
\newblock Lipschitz-constrained unsupervised skill discovery.
\newblock {\em arXiv preprint arXiv:2202.00914}, 2022.

\bibitem[\protect\citeauthoryear{Pathak \bgroup \em et al.\egroup }{2017}]{pathak2017curiosity}
Deepak Pathak, Pulkit Agrawal, Alexei~A Efros, and Trevor Darrell.
\newblock Curiosity-driven exploration by self-supervised prediction.
\newblock In {\em International conference on machine learning}, pages 2778--2787. PMLR, 2017.

\bibitem[\protect\citeauthoryear{Pathak \bgroup \em et al.\egroup }{2019}]{pathak2019self}
Deepak Pathak, Dhiraj Gandhi, and Abhinav Gupta.
\newblock Self-supervised exploration via disagreement.
\newblock In {\em International conference on machine learning}, pages 5062--5071. PMLR, 2019.

\bibitem[\protect\citeauthoryear{Raileanu and Rockt{\"a}schel}{2020}]{raileanu2020ride}
Roberta Raileanu and Tim Rockt{\"a}schel.
\newblock Ride: Rewarding impact-driven exploration for procedurally-generated environments.
\newblock In {\em International Conference on Learning Representations}, 2020.

\bibitem[\protect\citeauthoryear{Samvelyan \bgroup \em et al.\egroup }{2021}]{samvelyan2021minihack}
Mikayel Samvelyan, Robert Kirk, Vitaly Kurin, Jack Parker-Holder, Minqi Jiang, Eric Hambro, Fabio Petroni, Heinrich Kuttler, Edward Grefenstette, and Tim Rockt{\"a}schel.
\newblock Minihack the planet: A sandbox for open-ended reinforcement learning research.
\newblock In {\em Thirty-fifth Conference on Neural Information Processing Systems Datasets and Benchmarks Track (Round 1)}, 2021.

\bibitem[\protect\citeauthoryear{Schulman \bgroup \em et al.\egroup }{2017}]{schulman2017proximal}
John Schulman, Filip Wolski, Prafulla Dhariwal, Alec Radford, and Oleg Klimov.
\newblock Proximal policy optimization algorithms.
\newblock {\em arXiv preprint arXiv:1707.06347}, 2017.

\bibitem[\protect\citeauthoryear{Seo \bgroup \em et al.\egroup }{2021}]{seo2021state}
Younggyo Seo, Lili Chen, Jinwoo Shin, Honglak Lee, Pieter Abbeel, and Kimin Lee.
\newblock State entropy maximization with random encoders for efficient exploration.
\newblock In {\em Proceedings of the 38th International Conference on Machine Learning}, pages 9443--9454, 2021.

\bibitem[\protect\citeauthoryear{Stadie \bgroup \em et al.\egroup }{2015}]{stadie2015incentivizing}
Bradly~C Stadie, Sergey Levine, and Pieter Abbeel.
\newblock Incentivizing exploration in reinforcement learning with deep predictive models.
\newblock {\em arXiv preprint arXiv:1507.00814}, 2015.

\bibitem[\protect\citeauthoryear{Sutton and Barto}{2018}]{sutton2018reinforcement}
Richard~S Sutton and Andrew~G Barto.
\newblock {\em Reinforcement learning: An introduction}.
\newblock MIT press, 2018.

\bibitem[\protect\citeauthoryear{Tang \bgroup \em et al.\egroup }{2017}]{tang2017exploration}
Haoran Tang, Rein Houthooft, Davis Foote, Adam Stooke, OpenAI Xi~Chen, Yan Duan, John Schulman, Filip DeTurck, and Pieter Abbeel.
\newblock \# exploration: A study of count-based exploration for deep reinforcement learning.
\newblock {\em Advances in neural information processing systems}, 30, 2017.

\bibitem[\protect\citeauthoryear{Yarats \bgroup \em et al.\egroup }{2021}]{yarats2021reinforcement}
Denis Yarats, Rob Fergus, Alessandro Lazaric, and Lerrel Pinto.
\newblock Reinforcement learning with prototypical representations.
\newblock In {\em International Conference on Machine Learning}, pages 11920--11931. PMLR, 2021.

\bibitem[\protect\citeauthoryear{Yuan \bgroup \em et al.\egroup }{2022a}]{yuan2022rewarding}
Mingqi Yuan, Bo~Li, Xin Jin, and Wenjun Zeng.
\newblock Rewarding episodic visitation discrepancy for exploration in reinforcement learning.
\newblock In {\em Deep Reinforcement Learning Workshop NeurIPS 2022}, 2022.

\bibitem[\protect\citeauthoryear{Yuan \bgroup \em et al.\egroup }{2022b}]{yuan2022renyi}
Mingqi Yuan, Man-On Pun, and Dong Wang.
\newblock R{\'e}nyi state entropy maximization for exploration acceleration in reinforcement learning.
\newblock {\em IEEE Transactions on Artificial Intelligence}, 2022.

\bibitem[\protect\citeauthoryear{Yuan \bgroup \em et al.\egroup }{2024}]{yuan2024rlexplore}
Mingqi Yuan, Roger~Creus Castanyer, Bo~Li, Xin Jin, Glen Berseth, and Wenjun Zeng.
\newblock Rlexplore: Accelerating research in intrinsically-motivated reinforcement learning.
\newblock {\em arXiv preprint arXiv:2405.19548}, 2024.

\bibitem[\protect\citeauthoryear{Yuan \bgroup \em et al.\egroup }{2025}]{yuan2023rllte}
Mingqi Yuan, Zequn Zhang, Yang Xu, Shihao Luo, Bo~Li, Xin Jin, and Wenjun Zeng.
\newblock Rllte: Long-term evolution project of reinforcement learning.
\newblock In {\em Proceedings of the AAAI Conference on Artificial Intelligence}, 2025.

\bibitem[\protect\citeauthoryear{Zhang \bgroup \em et al.\egroup }{2021}]{zhang2021noveld}
Tianjun Zhang, Huazhe Xu, Xiaolong Wang, Yi~Wu, Kurt Keutzer, Joseph~E Gonzalez, and Yuandong Tian.
\newblock Noveld: A simple yet effective exploration criterion.
\newblock {\em Advances in Neural Information Processing Systems}, 34:25217--25230, 2021.

\end{thebibliography}
\clearpage\newpage

\appendix
\onecolumn
\section{Algorithmic Baselines}\label{appendix:baselines}
 
 \textbf{ICM} \cite{pathak2017curiosity}. ICM leverages an inverse-forward model to learn the dynamics of the environment and uses the prediction error as the curiosity reward. Specifically, the inverse model inferences the current action $\bm{a}_{t}$ based on the encoded states $\bm{e}_{t}$ and $\bm{e}_{t+1}$, where $\bm{e}=\psi(\bm{s})$ and $\psi(\cdot)$ is an embedding network. Meanwhile, the forward model $f$ predicts the encoded next-state $\bm{e}_{t}$ based on $(\bm{e}_{t},\bm{a}_t)$. Finally, the intrinsic reward is defined as
 \begin{equation}
     I_{t}=\Vert f(\bm{e}_{t},\bm{a}_{t})-\bm{e}_{t+1}\Vert_{2}^{2}.
 \end{equation}

\noindent\textbf{NGU} \cite{badia2020never}. NGU is a mixed intrinsic reward approach that combines global and episodic exploration and the first algorithm to achieve non-zero rewards in the game of \textit{Pitfall!} without using demonstrations or hand-crafted features. The intrinsic reward is defined as
 \begin{equation}
     I_{t}=\min\{\max\{\alpha_{t}\}, C\}/\sqrt{N_{\rm ep}(\bm{s}_{t})},
 \end{equation}
 where $\alpha_{t}$ is a life-long curiosity factor computed following the RND method, $C$ is a chosen maximum reward scaling, and $N_{\rm ep}$ is the episodic state visitation frequency computed by pseudo-counts. More specifically, $N_{\rm ep}$ is computed as
 \begin{equation}
     \sqrt{N_{\rm ep}(\bm{s}_{t})}\approx\sqrt{\sum_{\tilde{\bm{e}}_{i}}K(\tilde{\bm{e}}_{i},\bm{e}_t)}+c,
 \end{equation}
 where $\tilde{\bm{e}}_{i}$ is the first $k$ nearest neighbors of $\bm{e}$, $K$ is a Dirac delta function, and $c$ guarantees a minimum amount of pseudo-counts. 

\noindent\textbf{RE3} \cite{seo2021state}. RE3 is an information theory-based and computation-efficient exploration approach that aims to maximize the Shannon entropy of the state visiting distribution. In particular, RE3 leverages a random and fixed neural network to encode the state space and employs a $k$-nearest neighbor estimator to estimate the entropy efficiently. Then, the estimated entropy is transformed into particle-based intrinsic rewards. Specifically, the intrinsic reward is defined as
 \begin{equation}
     I_{t}=\frac{1}{k}\sum_{i=1}^{k}\log(\Vert\bm{e}_{t}-\tilde{\bm{e}}_{t}^{i}\Vert_{2}+1).
 \end{equation}

\noindent\textbf{E3B} \cite{henaff2022exploration}. E3B provides a generalization of count-based rewards to continuous spaces. E3B learns a representation mapping from observations to a latent space (e.g., using inverse dynamics). At each episode, the sequence of latent observations parameterizes an ellipsoid \cite{li2010contextual,auer2002using,dani2008stochastic}, which is used to measure the novelty of the subsequent observations. In tabular settings, the E3B ellipsoid reduces to the table of inverse state-visitation frequencies \cite{henaff2022exploration}. Given a feature encoding $f$, at each time step $t$ of the episode the elliptical bonus $I_{t}$ is defined as follows: 
 
 \begin{equation}
     I_{t}=f(\bm{s}_{t})^T C_{t-1} f(\bm{s}_{t}),
 \end{equation}

  \begin{equation}
     C_{t-1} = \sum_{i=1}^{t-1} f(\bm{s}_{i})f(\bm{s}_{i})^T + \lambda \mathbf{I},
 \end{equation}

 where $f$ is the learned representation mapping, $C_{t-1}$ is the episodic ellipsoid \cite{henaff2022exploration}, $\lambda$ is a scalar
coefficient, and $\mathbf{I}$ is the identity matrix.

\clearpage\newpage

\section{Experimental Settings}\label{appendix:exp setting}

\subsection{Baselines}
In this paper, we utilize the implementations provided in \cite{yuan2023rllte,yuan2024rlexplore} for the baseline intrinsic rewards. In particular, \cite{yuan2024rlexplore} examines how low-level implementation details affect the performance of intrinsic rewards. Therefore, we follow the recommended configuration for these baseline intrinsic rewards in our experiments, as detailed in Table~\ref{tb:bs irs}. Note that these configurations remain fixed for all the experiments.

\begin{table}[h!]
\centering
\caption{Configuration of the baseline intrinsic rewards. Here, \textit{RMS} refers to the use of an exponential moving average of the mean and standard deviation for normalization.}
\vskip 0.15in
\label{tb:bs irs}
\begin{tabular}{lllll}
\toprule
\textbf{Hyperparameter}   & \textbf{ICM} & \textbf{NGU} & \textbf{RE3} & \textbf{E3B} \\ \midrule
Observation normalization & Min-Max      & RMS          & RMS          & RMS          \\
Reward normalization      & RMS          & RMS          & Min-Max      & RMS          \\
Weight initialization     & Orthogonal   & Orthogonal   & Orthogonal   & Orthogonal   \\
Update proportion         & 1.0          & 1.0          & N/A          & 1.0          \\
with LSTM                 & False        & False        & False        & False        \\ \bottomrule
\end{tabular}
\vskip -0.1in
\end{table}

The initial exploration coefficient $\beta_{0}$ is critical for all the experiments. Therefore, we did a grid search for $\beta_{0}\in[0.1, 0.25, 0.5, 1.0]$ and found the best values are 0.25 for MiniGrid, 0.1 for Procgen, and 0.1 for ALE-5, which were used to produce all the results in this paper.

\subsection{Backbone RL Algorithm}
The PPO serves as the backbone RL algorithm, and Table~\ref{tb:ppo_params} illustrates the detailed hyperparameters, which also remain fixed for all the experiments.
\begin{table}[!h]
\centering
\caption{PPO hyperparameters for MiniGrid, Procgen, and ALE-5.}
\vskip 0.15in
\label{tb:ppo_params}
\begin{tabular}{llll}
\toprule
\textbf{Hyperparameter}    & \textbf{ALE-5} & \textbf{MiniGrid} & \textbf{Procgen} \\ \midrule
Observation downsampling   & (84, 84)                & (7, 7)            & (64, 64)         \\
Observation normalization  & / 255.                  & No                & / 255.           \\
Reward normalization       & No                      & No                & No               \\
Weight initialization      & Orthogonal              & Orthogonal        & Orthogonal       \\
LSTM                       & No                      & No                & No               \\
Stacked frames             & 4                      & No                & No               \\
Pre-training steps         & 5000000  & N/A  & N/A \\
Environment steps          & 5000000                & 10000000          & 25000000  
\\
Episode steps              & 128                     & 32                & 256              \\
Number of workers          & 1                       & 1                 & 1                \\
Environments per worker    & 8                       & 256               & 64               \\
Optimizer                  & Adam                    & Adam              & Adam             \\
Learning rate              & 2.5e-4                  & 2.5e-4            & 5e-4             \\
GAE coefficient            & 0.95                    & 0.95              & 0.95             \\
Action entropy coefficient & 0.01                    & 0.01              & 0.01             \\
Value loss coefficient     & 0.5                     & 0.5               & 0.5              \\
Value clip range           & 0.1                     & 0.1               & 0.2              \\
Max gradient norm          & 0.5                     & 0.5               & 0.5              \\
Epochs per rollout         & 4                       & 4                 & 3                \\
Batch size                 & 256                     & 1024              & 2048             \\
Discount factor            & 0.99                    & 0.99              & 0.999            \\ \bottomrule
\end{tabular}
\vskip -0.1in
\end{table}

\clearpage\newpage

\section{Performance Rankings}\label{appendix:ranking}
\subsection{Rankings}
\subsubsection{MiniGrid}
\begin{figure*}[h!]
    \centering
    \includegraphics[width=0.94\linewidth]{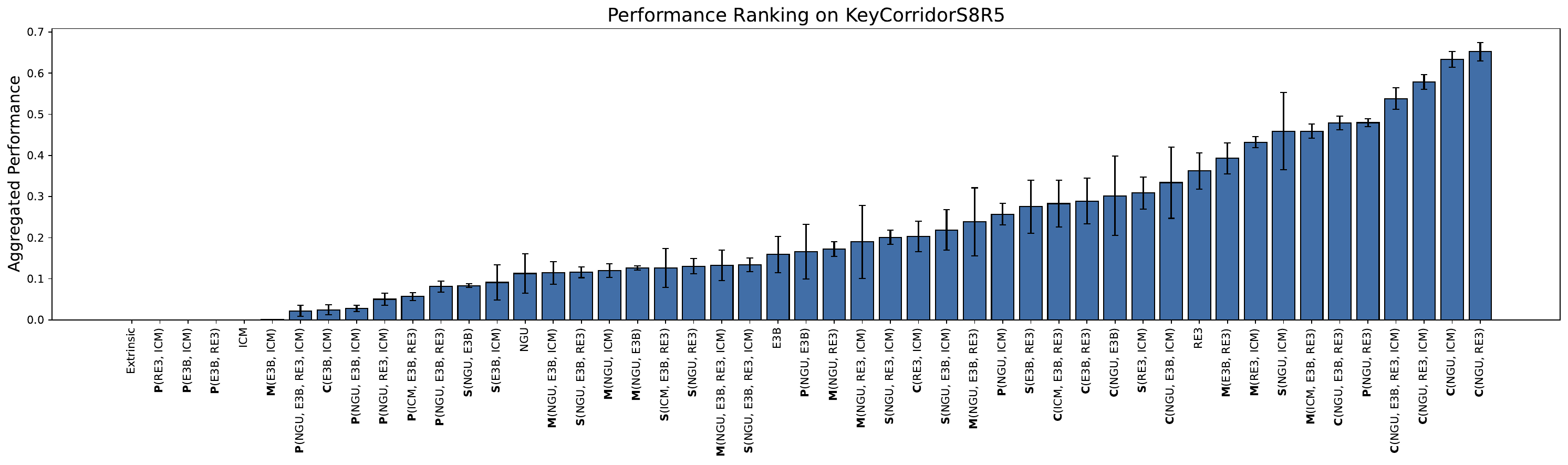}
    \includegraphics[width=0.94\linewidth]{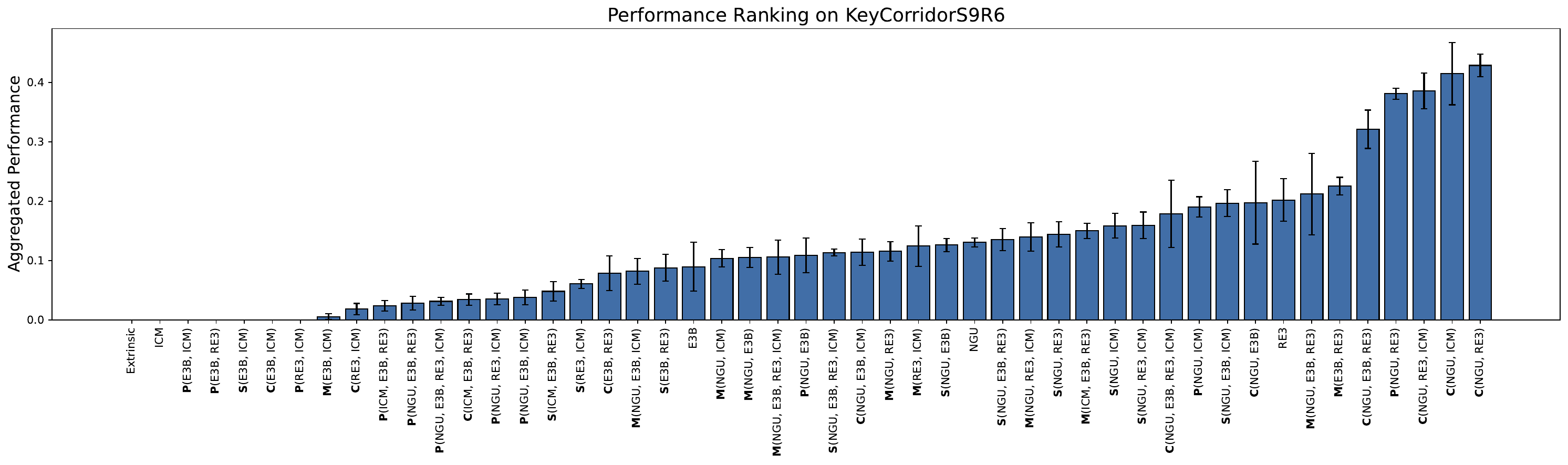}
    \includegraphics[width=0.94\linewidth]{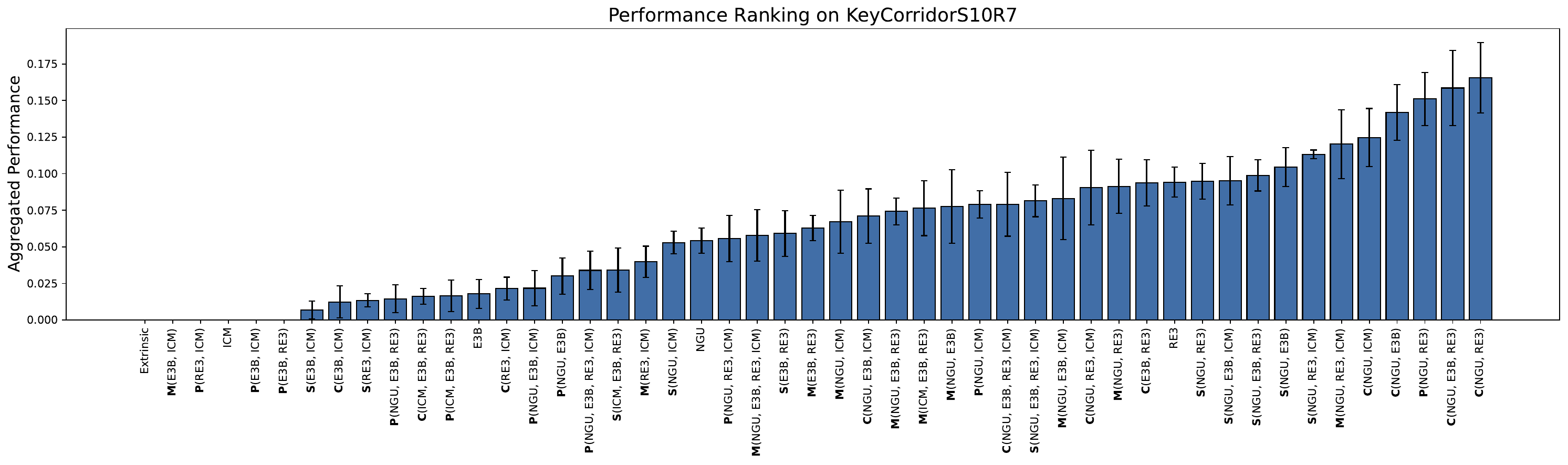}
    \includegraphics[width=0.94\linewidth]{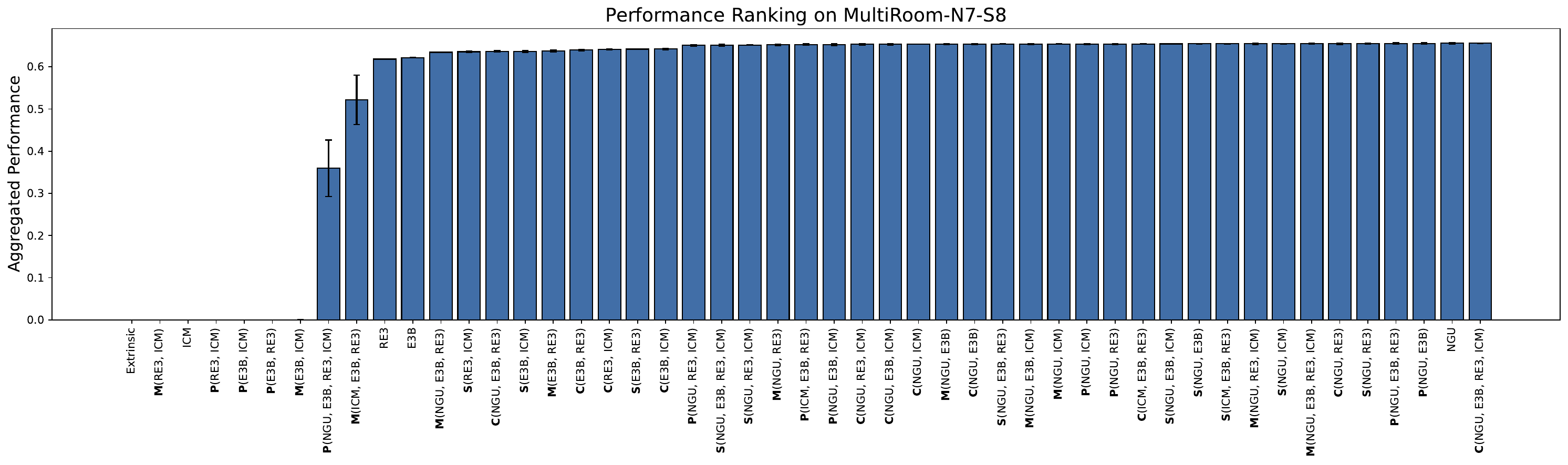}
    \caption{Performance ranking on \textit{KeyCorridorS8R5}, \textit{KeyCorridorS9R6}, \textit{KeyCorridorS10R7}, and \textit{MultiRoom-N7-S8}. The mean and standard error are computed using five random seeds.}
\end{figure*}

\clearpage\newpage

\begin{figure*}[h!]
    \centering
    \includegraphics[width=\linewidth]{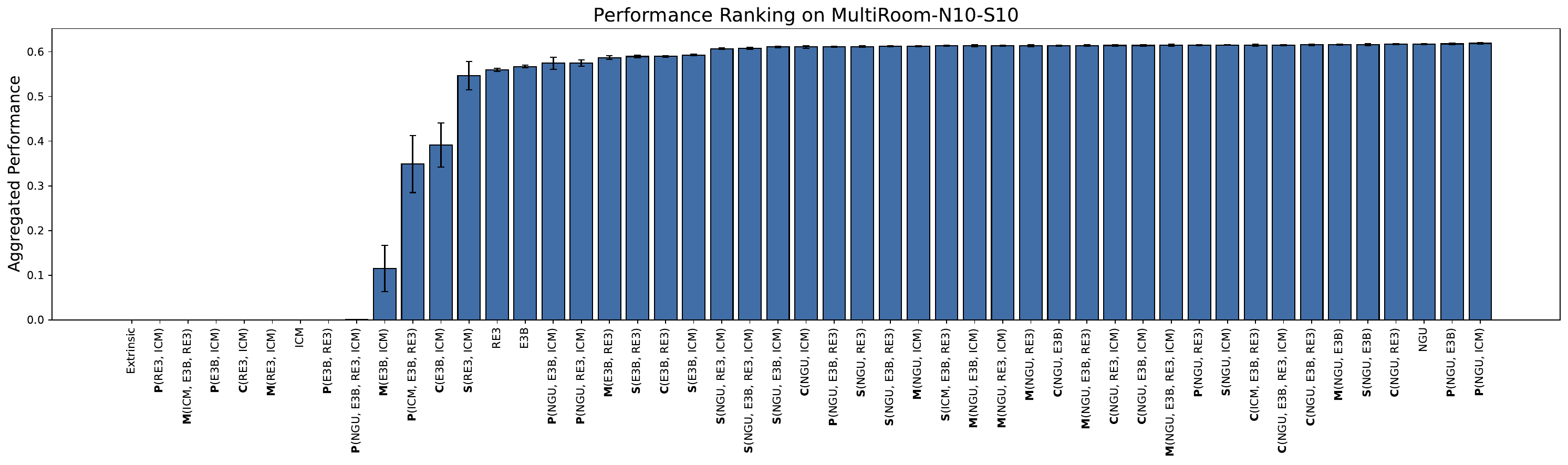}
    \includegraphics[width=\linewidth]{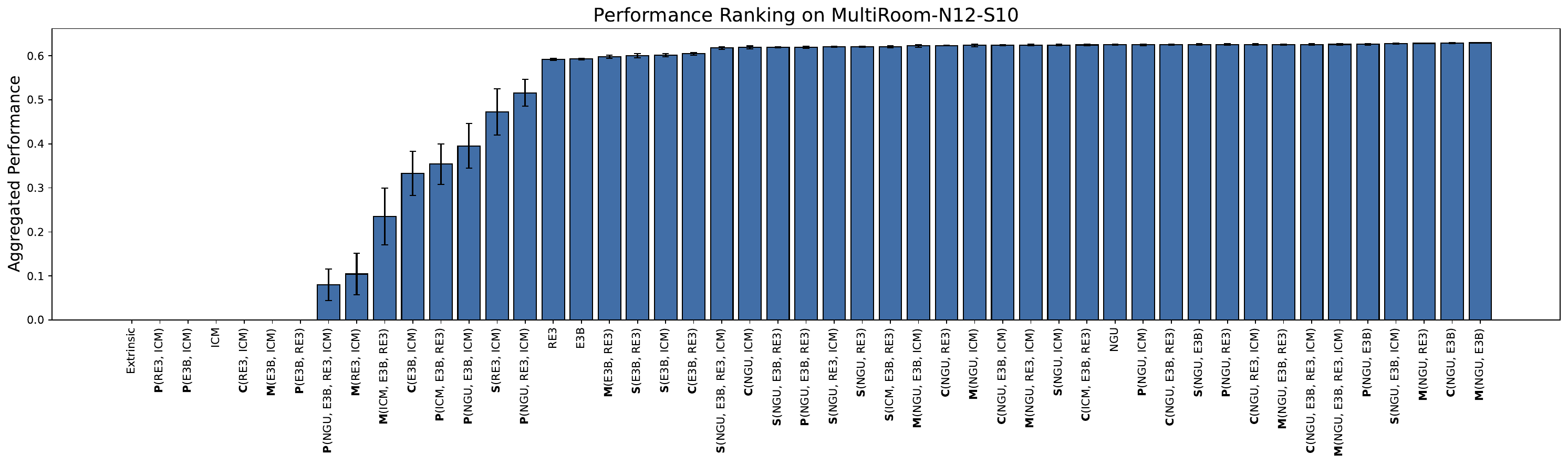}
    \includegraphics[width=\linewidth]{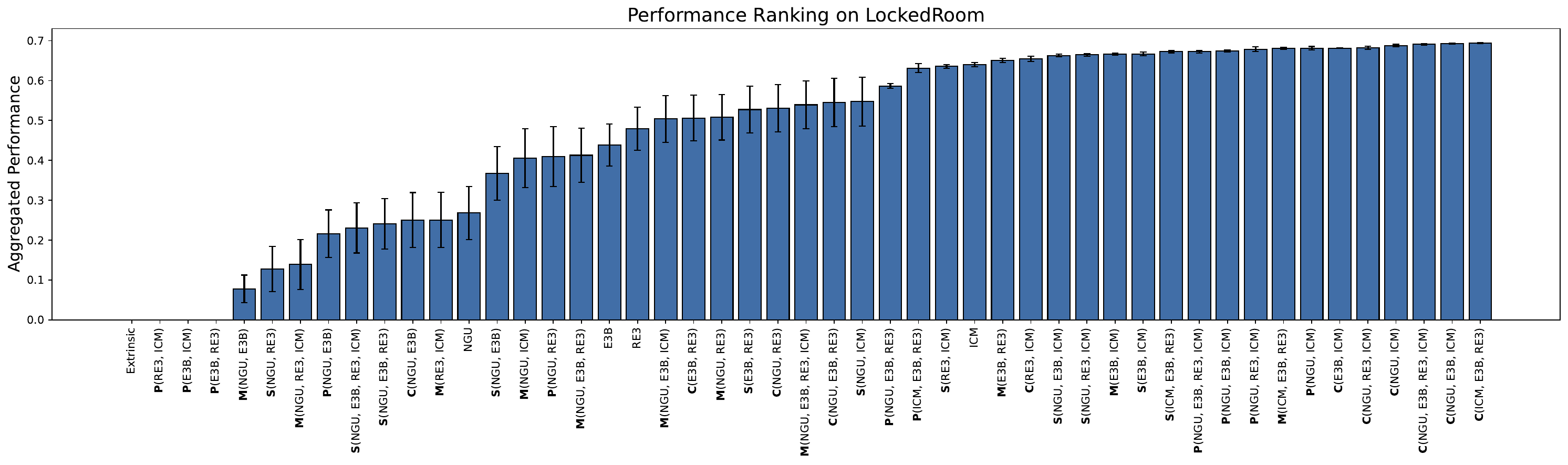}
    \includegraphics[width=\linewidth]{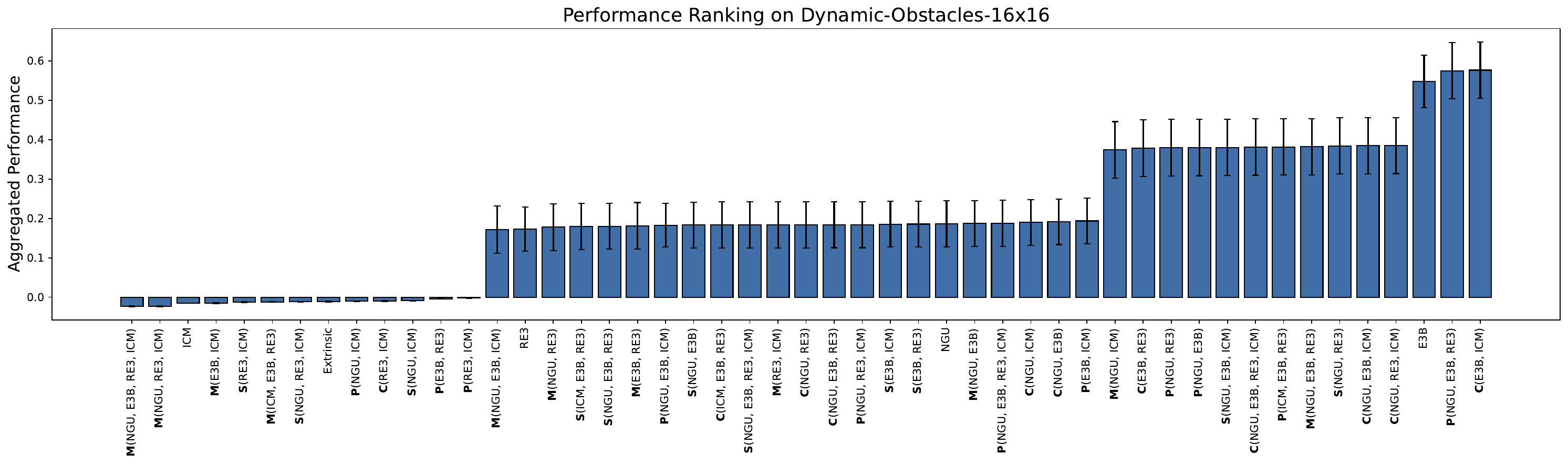}
    \caption{Performance ranking on \textit{MultiRoom-N10-S10}, \textit{MultiRoom-N12-S10}, \textit{LockedRoom}, and \textit{Dynamic-Obstacles-16×16}. The mean and standard error are computed using five random seeds.}
\end{figure*}

\clearpage\newpage

\subsubsection{Procgen}
\begin{figure}[h!]
    \centering
    \includegraphics[width=0.94\linewidth]{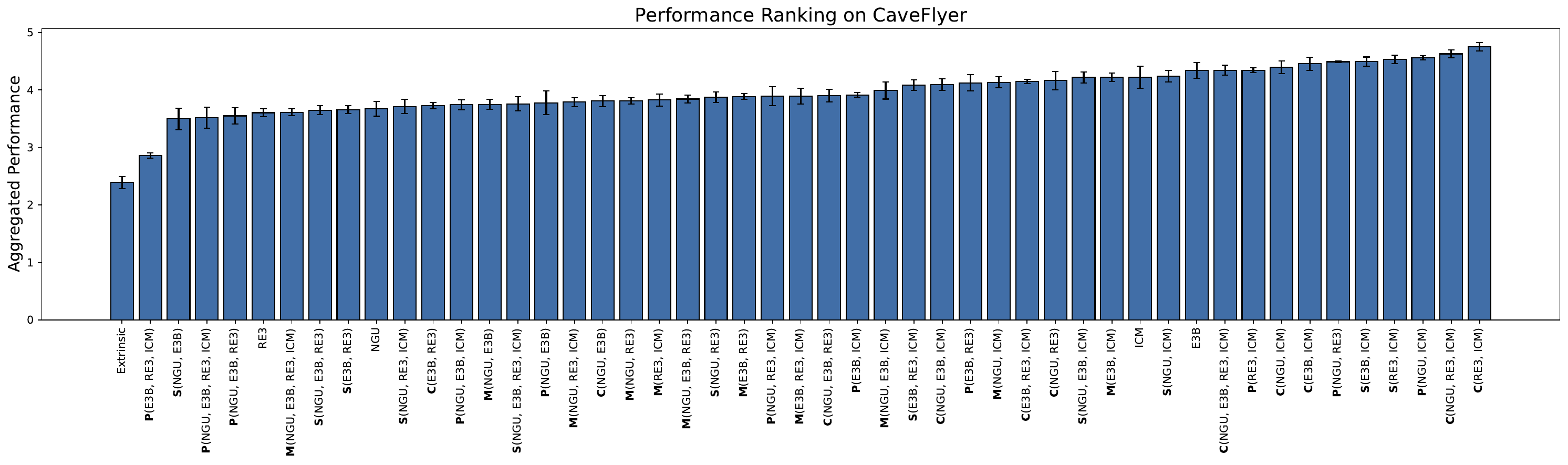}
    \includegraphics[width=0.94\linewidth]{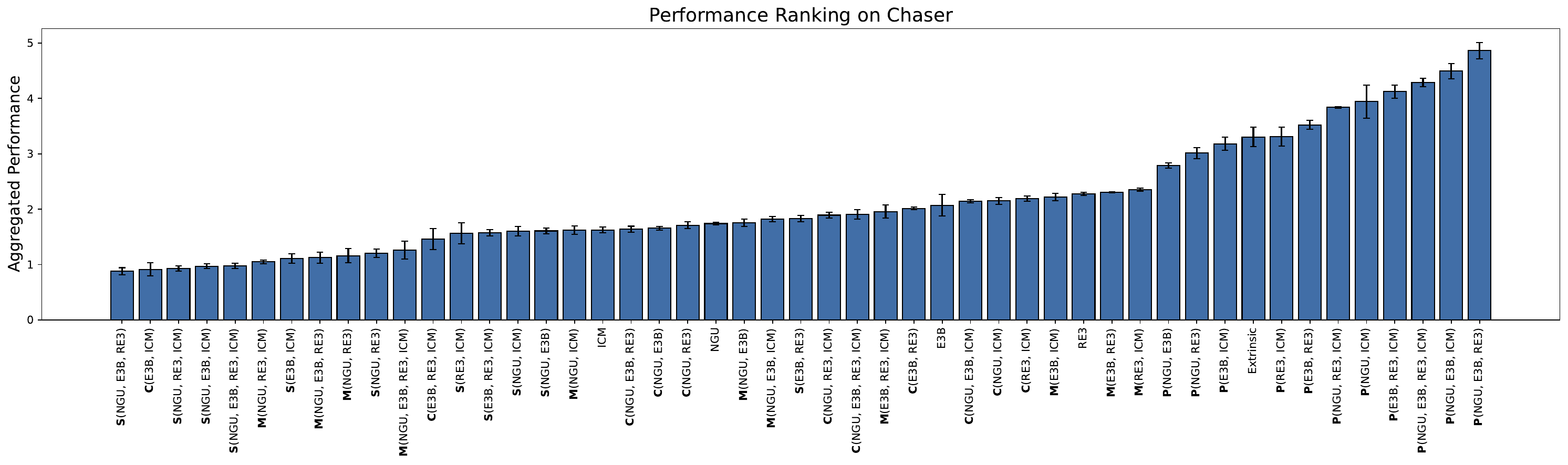}
    \includegraphics[width=0.94\linewidth]{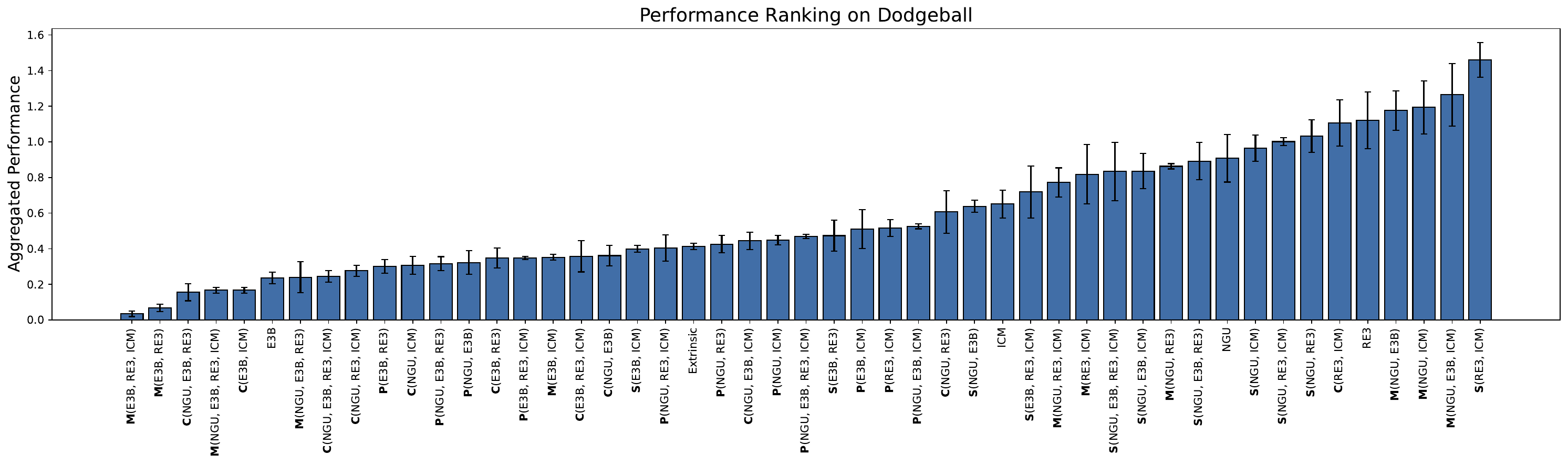}
    \includegraphics[width=0.94\linewidth]{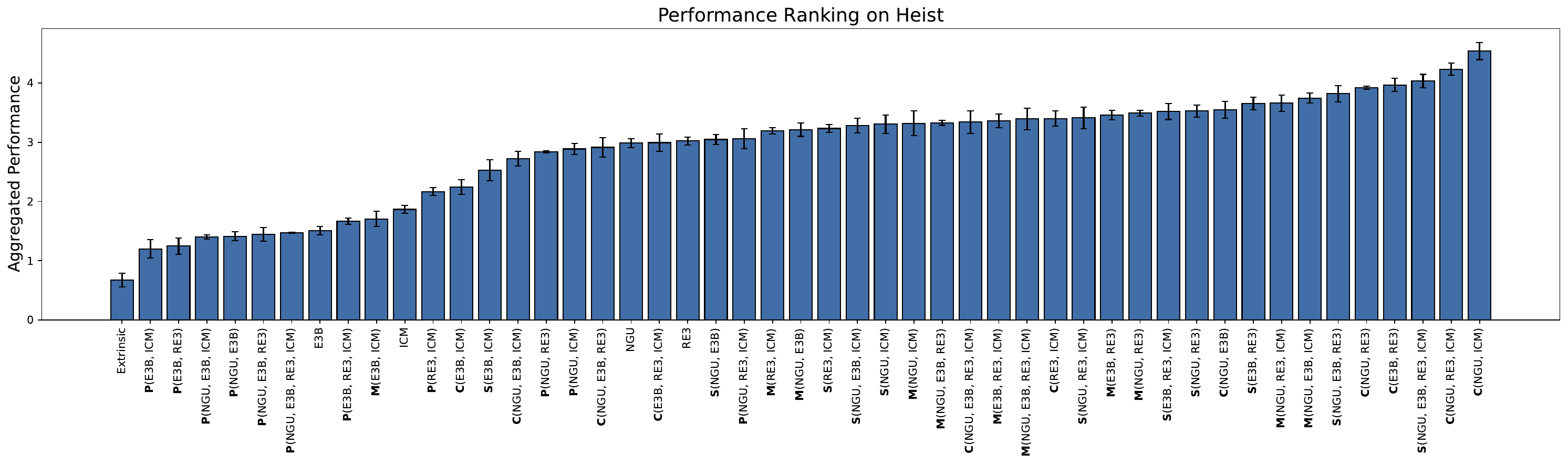}
    \caption{Performance ranking on \textit{CaveFlyer}, \textit{Chaser}, \textit{Dodgeball}, and \textit{Heist}. The mean and standard error are computed using five random seeds.}
\end{figure}

\clearpage\newpage

\begin{figure}[h!]
    \centering
    \includegraphics[width=\linewidth]{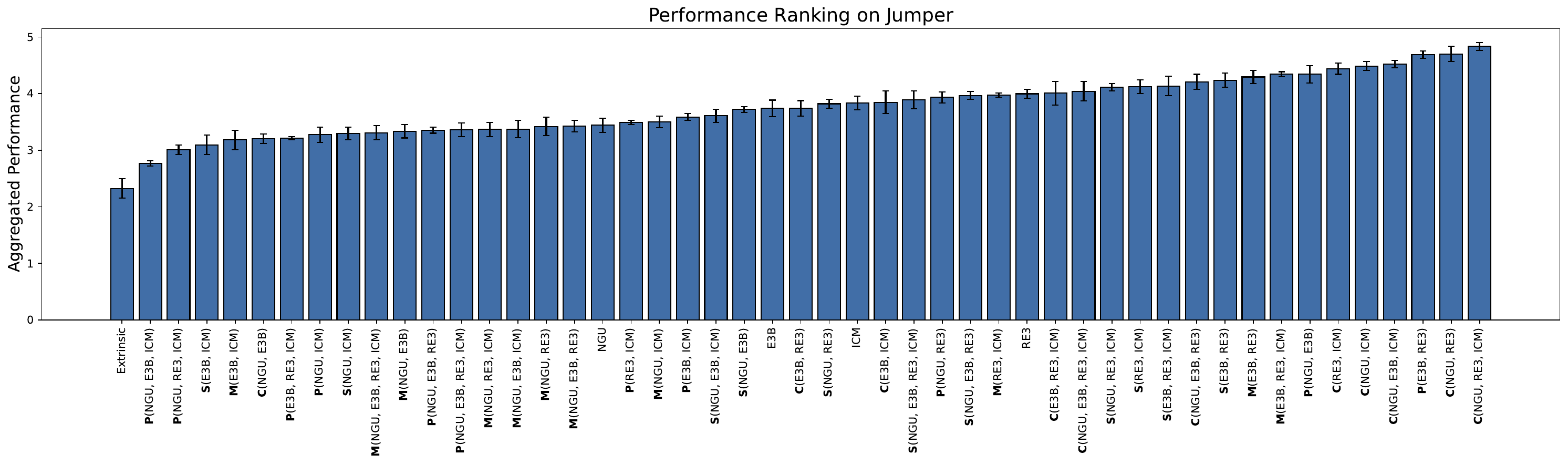}
    \includegraphics[width=\linewidth]{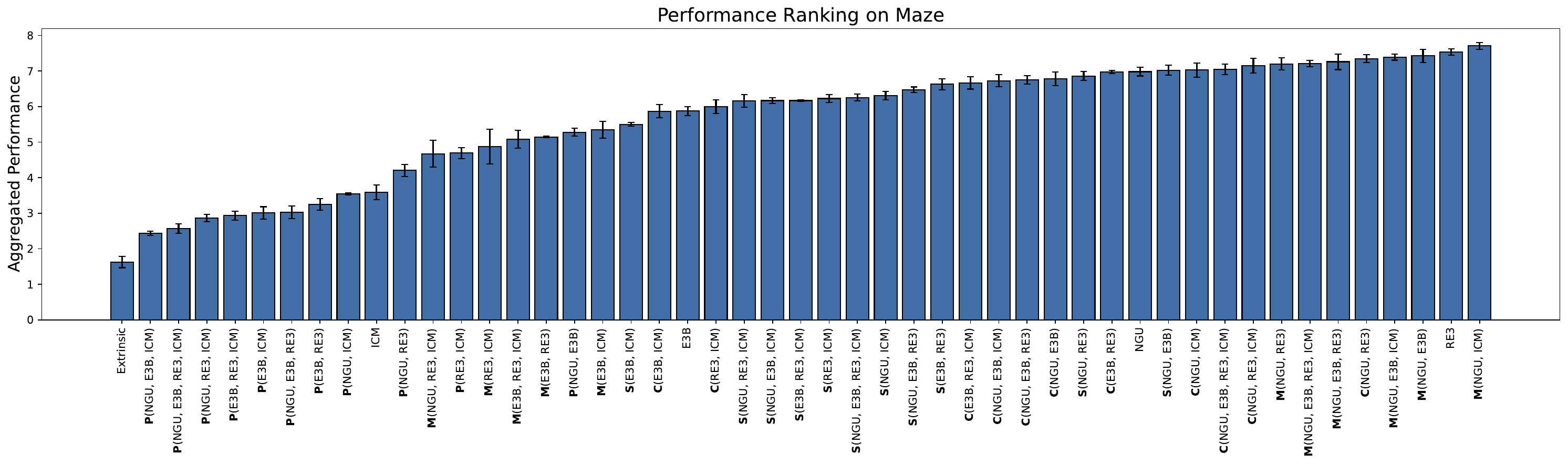}
    \includegraphics[width=\linewidth]{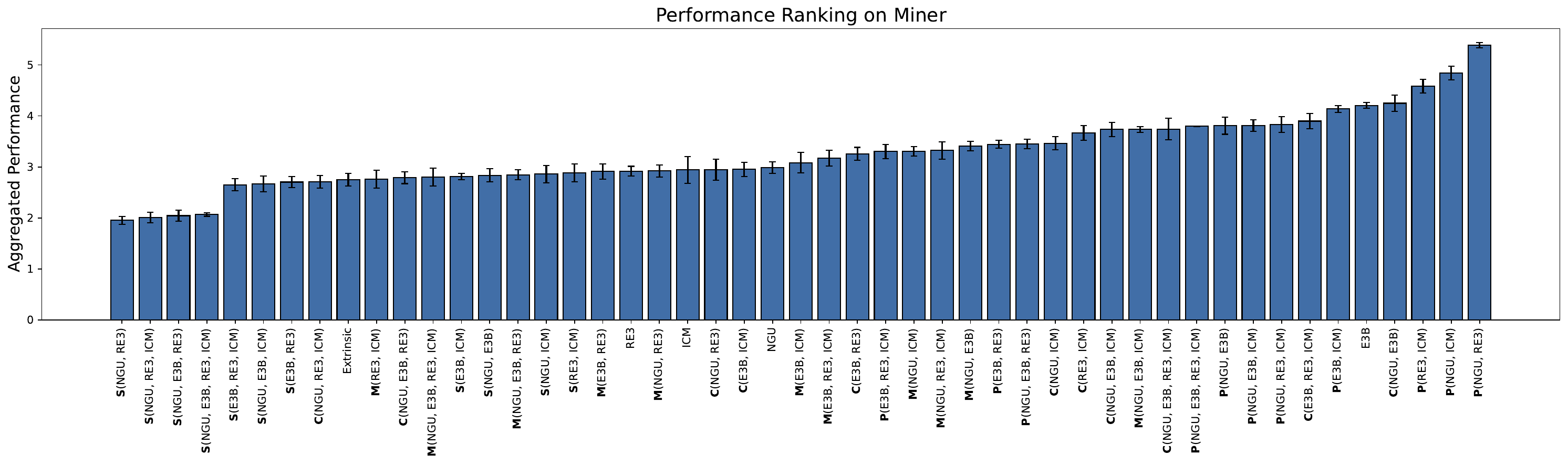}
    \includegraphics[width=\linewidth]{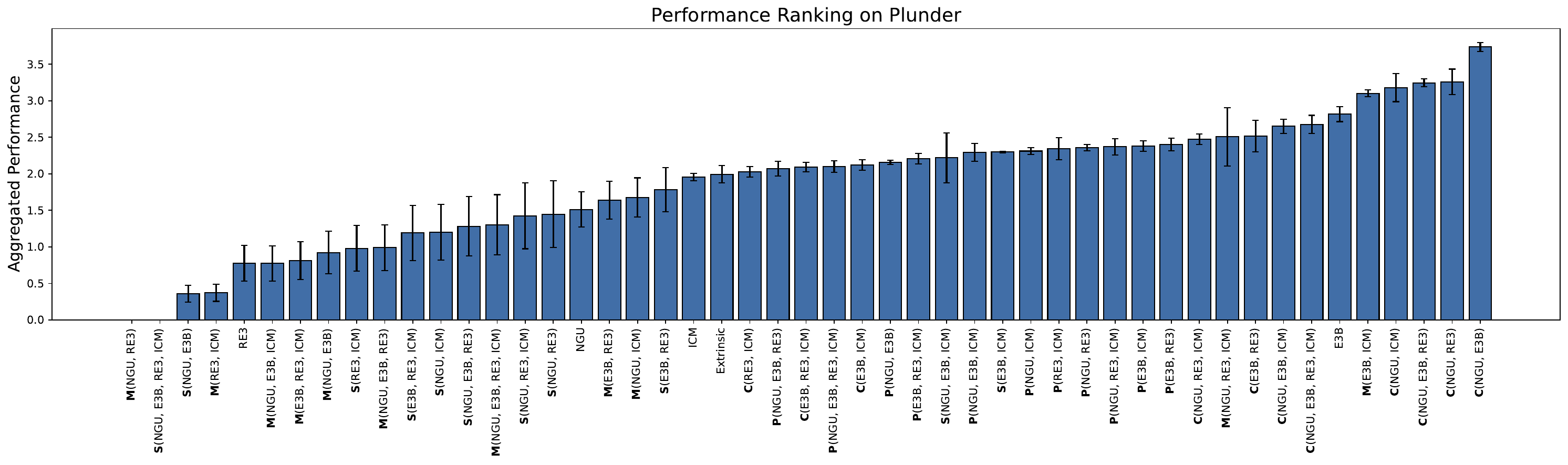}
    \caption{Performance ranking on \textit{Jumper}, \textit{Maze}, \textit{Miner}, and \textit{Plunder}. The mean and standard error are computed using five random seeds.}
\end{figure}

\clearpage\newpage

\subsubsection{ALE}

\begin{figure}[h!]
    \centering
    \includegraphics[width=0.78\linewidth]{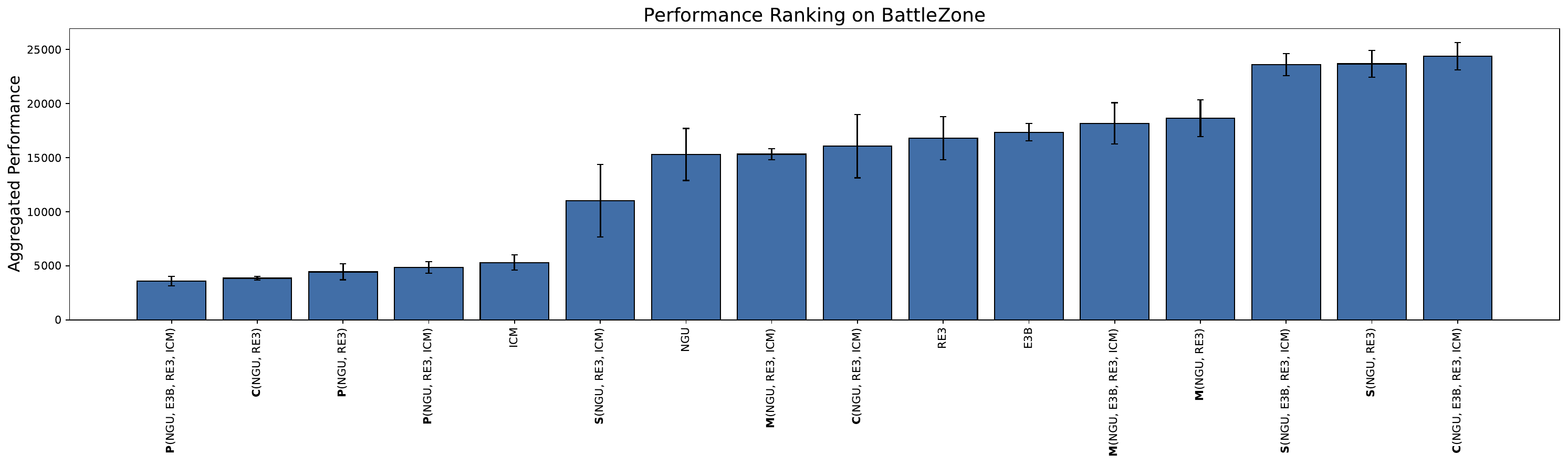}
    \includegraphics[width=0.78\linewidth]{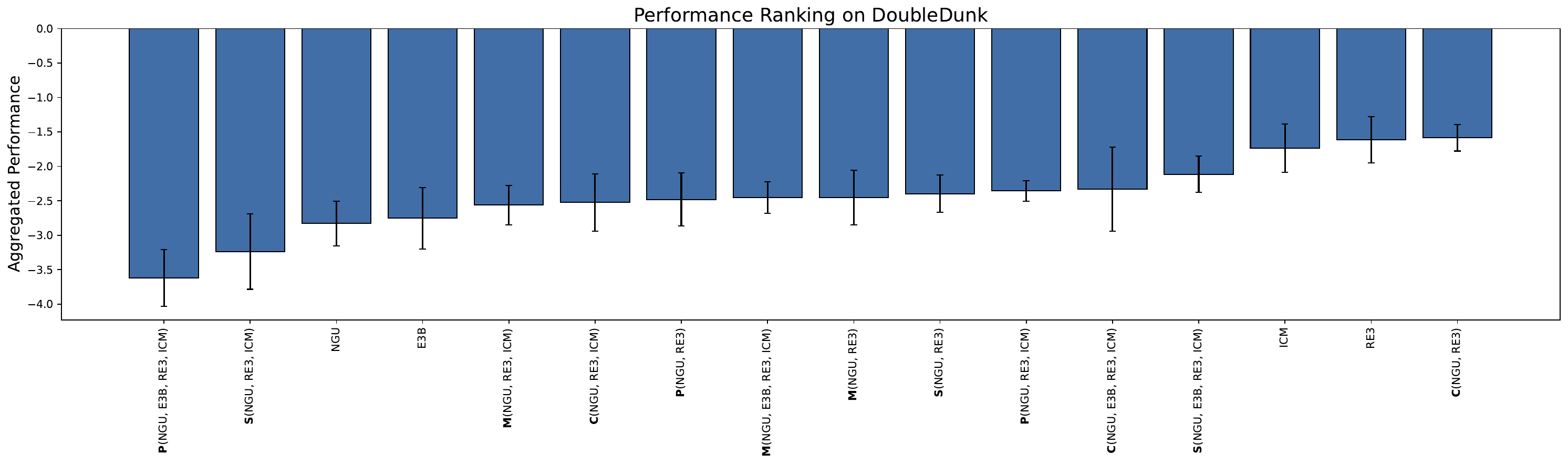}
    \includegraphics[width=0.78\linewidth]{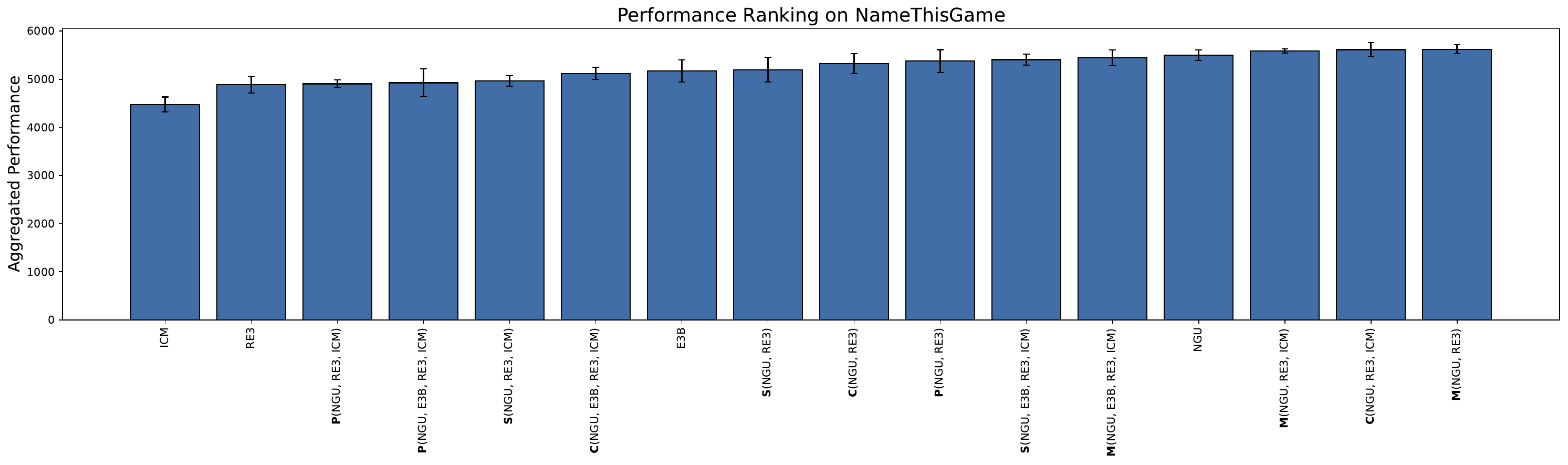}
    \includegraphics[width=0.78\linewidth]{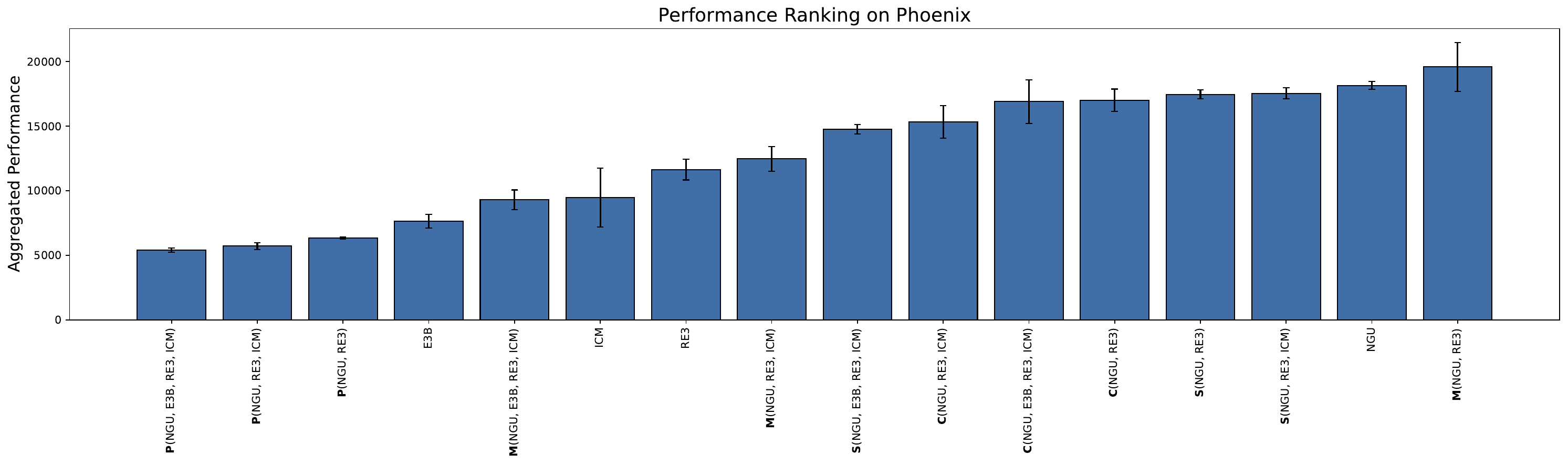}
    \includegraphics[width=0.78\linewidth]{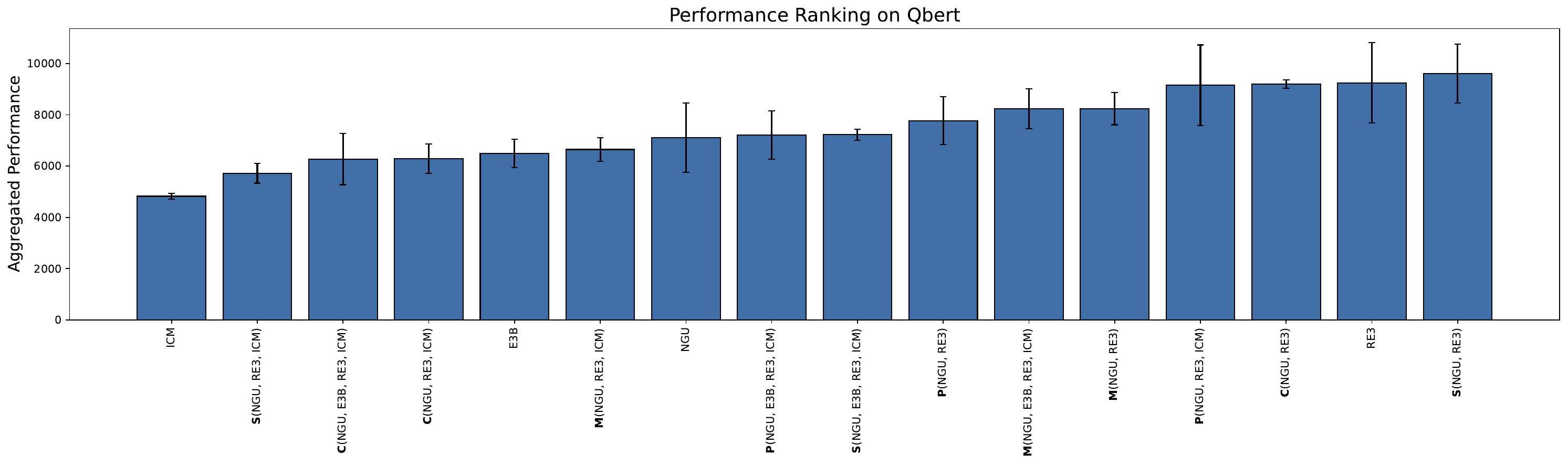}
    \caption{Performance ranking on \textit{BattleZone}, \textit{DoubleDunk}, \textit{NameThisGame}, \textit{Phoenix}, and \textit{Q*bert}. The mean and standard error are computed using five random seeds.}
\end{figure}

\clearpage\newpage

\subsection{Proportion of Top Candidates}

\begin{table}[h!]
\centering
\begin{tabular}{ccccccc}
\toprule[1.0pt]
\textbf{Strategy} & Extrinsic & Baseline & Summation & Product & Maximum & Cycle            \\ \midrule[1.0pt]
\textbf{Top 1}    & 0         & 0 & 0 & 12.50\% & 12.50\% & \textbf{75.00\%}  \\ \midrule
\textbf{Top 5}    & 0         & 7.50\% & 7.50\% & 22.50\% & 5.0\% & \textbf{57.50\%}  \\ \midrule
\textbf{Top 10}   & 0 & 5.00\% & 17.50\% & 17.50\% & 18.75\% & \textbf{41.25\%}  \\ \midrule
\textbf{Top 20}   & 0 & 6.25\% & 21.88\% & 16.25\% & 21.88\% & \textbf{33.75\%} \\ \bottomrule[1.0pt]
\end{tabular}
\caption{Proportion of each fusion strategy in the top reward candidates for each MiniGrid environment. The highest values are shown in bold.}
\label{tb:minigrid top}
\end{table}

\begin{table}[h!]
\centering
\begin{tabular}{ccccccc}
\toprule[1.0pt]
\textbf{Strategy} & Extrinsic & Baseline & Summation & Product & Maximum & Cycle            \\ \midrule[1.0pt]
\textbf{Top 1}    & 0         & 0        & 12.50\%    & 25.00\%  & 12.50\%  & \textbf{50.00\%}  \\ \midrule
\textbf{Top 5}    & 0 & 7.50\% & 10.00\% & 25.00\% & 17.50\% & \textbf{40.00\%}    \\ \midrule
\textbf{Top 10}   & 1.25\% & 6.25\% & 12.50\% & 26.25\% & 18.75\% & \textbf{35.00\%}  \\ \midrule
\textbf{Top 20}   & 0.62\% & 7.50\% & 18.12\% & 21.88\% & 20.00\% & \textbf{31.87\%} \\ \bottomrule[1.0pt]
\end{tabular}
\caption{Proportion of each fusion strategy in the top reward candidates for each Procgen environment. The highest values are shown in bold.}
\label{tb:procgen top}
\end{table}

\subsection{Best Reward Candidate for Each Environment}

\begin{table}[h!]
\centering
\begin{tabular}{llll}
\toprule[1.0pt]
\textbf{Environment}           & \textbf{Candidate}    & \textbf{Environment} & \textbf{Candidate}    \\ \midrule[1.0pt]
KeyCorridorS8R5         & C(NGU, RE3)           & CaveFlyer     & C(RE3, ICM) \\
KeyCorridorS9R6         & C(NGU, RE3)           & Chaser        & P(NGU, E3B, RE3)      \\
KeyCorridorS10R7        & C(NGU, RE3)           & Dodgeball     & S(RE3, ICM)           \\
MultiRoom-N7-S8         & C(NGU, E3B, RE3, ICM) & Heist         & C(NGU, ICM) \\
MultiRoom-N10-S10       & P(NGU, ICM)           & Jumper        & C(NGU, RE3, ICM)      \\
MultiRoom-N12-S10       & M(NGU, E3B)           & Maze          & M(NGU, ICM)           \\
LockedRoom              & C(E3B, RE3, ICM)      & Miner         & P(NGU, RE3)           \\
Dynamic-Obstacles-16×16 & C(E3B, ICM)           & Plunder       & C(NGU, E3B)           \\ \bottomrule[1.0pt]
\end{tabular}
\caption{Best reward candidates for MiniGrid and Procgen environments.}
\label{tb:best_mg_procgen}
\end{table}

\begin{table}[h!]
\centering
\begin{tabular}{llll}
\toprule[1.0pt]
\textbf{Environment}           & \textbf{Candidate}  \\ \midrule[1.0pt]
BattleZone         & C(NGU, E3B, RE3, ICM) \\
DoubleDunk         & C(NGU, RE3)      \\
NameThisGame        & M(NGU, RE3)           \\
Phoenix        & M(NGU, RE3) \\
Q*bert      & S(NGU, RE3)         \\ \bottomrule[1.0pt]
\end{tabular}
\caption{Best reward candidates for the ALE-5 benchmark.}
\label{tb:best_ale}
\end{table}

\clearpage\newpage

\section{Quantity-level Performance Comparison}

\begin{figure*}[h!]
    \centering
    \vskip 0.2in
    \includegraphics[width=\linewidth]{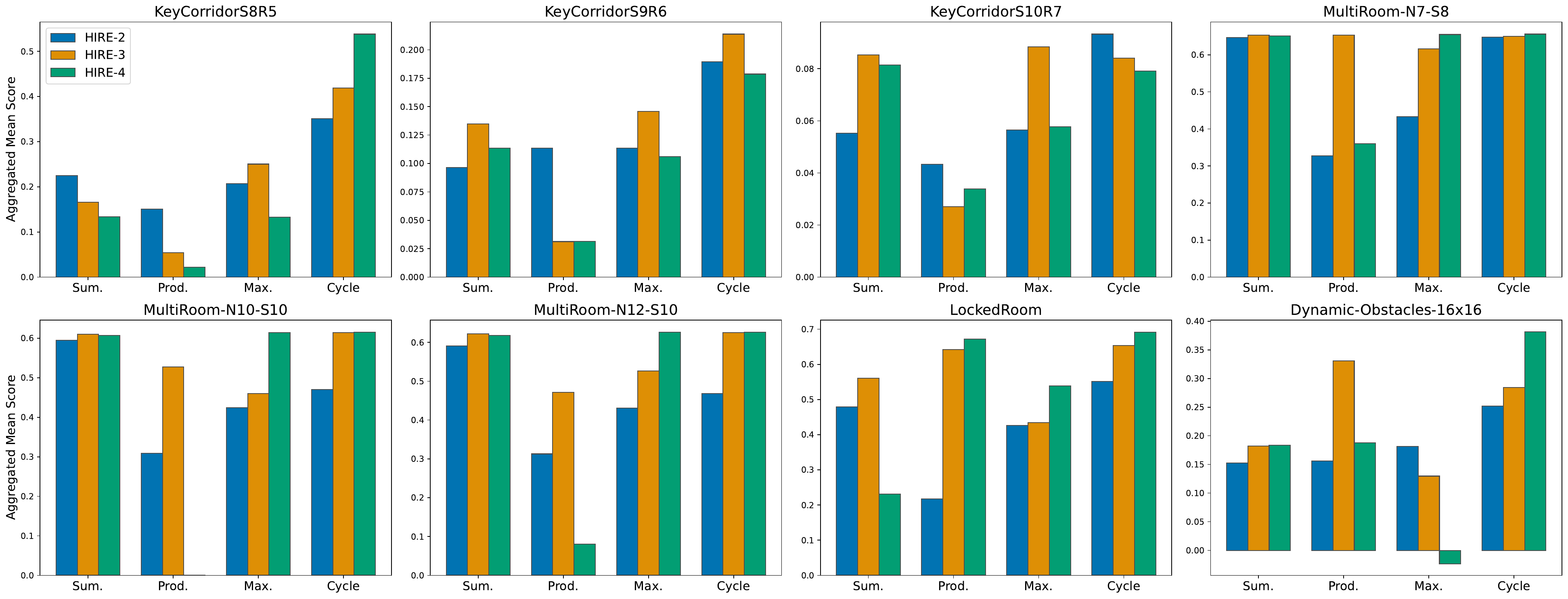}
    \caption{Quantity-level performance comparison on the MiniGrid benchmark.}
    \label{fig:minigrid_number_level_single}
    \vskip -0.2in
\end{figure*}

\begin{figure*}[h!]
    \centering
    \vskip 0.2in
    \includegraphics[width=\linewidth]{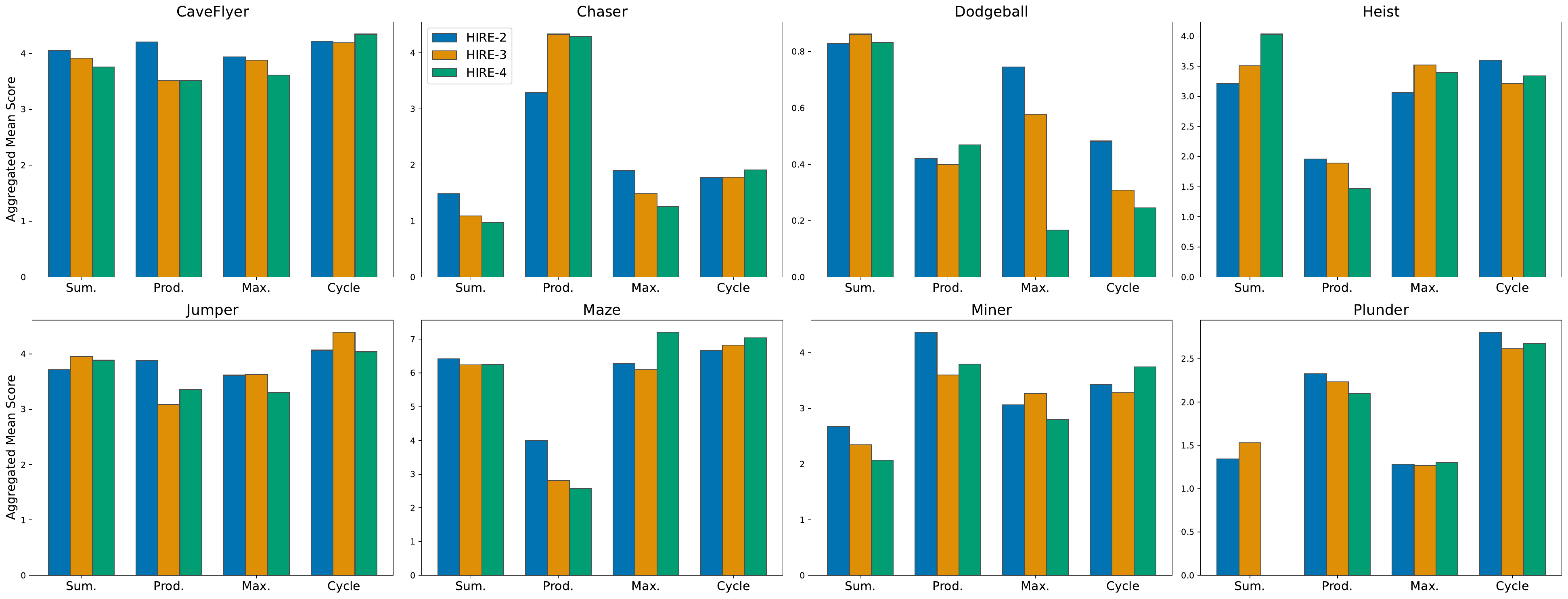}
    \caption{Quantity-level performance comparison on the Procgen benchmark.}
    \label{fig:procgen_number_level_single}
    \vskip -0.2in
\end{figure*}

\clearpage\newpage

\end{document}